\documentclass[11pt,letterpaper]{article}
\usepackage{emnlp2017}
\usepackage{times}
\usepackage{latexsym}
\usepackage{amsmath}
\usepackage{url}
\usepackage{booktabs, adjustbox}
\usepackage{longtable}
\usepackage{float}
\usepackage{multirow}
\usepackage{subfigure}
\usepackage{color}

\emnlpfinalcopy

\def\emnlppaperid{906}

\newcommand{\visual}{\textsc{vis}}
\newcommand{\textual}{\textsc{tex}}
\newcommand{\combi}{\textsc{combi}}
\newcommand{\visualPca}{\textsc{visPca}}
\newcommand{\combiPca}{\textsc{combiPca}}

\title{Limitations of Cross-Lingual Learning from Image Search}

\author{Mareike Hartmann \\
  University of Copenhagen\\
  {\tt hartmann@di.ku.dk} \\\And
  Anders S\o gaard \\
  University of Copenhagen \\
  {\tt  soegaard@di.ku.dk} \\}

\date{}

\begin{document}

\maketitle

\begin{abstract} 
Cross-lingual representation learning is an important step in making NLP scale to all the world's languages. Recent work on bilingual lexicon induction suggests that it is possible to learn cross-lingual representations of words based on similarities between images associated with these words. However, that work focused on the translation of selected nouns only. In our work, we investigate whether the meaning of other parts-of-speech, in particular adjectives and verbs, can be learned in the same way. We also experiment with combining the representations learned from visual data with embeddings learned from textual data. Our experiments across five language pairs indicate that previous work does not scale to the problem of learning cross-lingual representations beyond simple nouns. 
\end{abstract}

\section{Introduction}
Typically, cross-lingual word representations are learned from word alignments, sentence alignments, or, more rarely, from aligned, comparable documents \cite{Levy:ea:17}. \citet{Ammar2016} propose a method that learns multilingual word-embeddings from monolingual corpora and dictionaries containing translation pairs. However, for many languages such resources are not available. 

\citet{Bergsma2011Vis} introduced an alternative idea, namely to learn bilingual representations from image data collected via web image search. The idea behind their approach is to represent words in a visual space and find valid translations between words based on similarities between their visual representations. Representations of words in the visual space are built by representing a word by a set of images that are associated with that word, i.e., the word is a semantic tag for the images in the set. 

\citet{Kiela2015} gain large performance improvements for the same task using a feature representation extracted from convolutional neural networks. However, both works only consider nouns, leaving open the question of whether learning cross-lingual representations for other parts-of-speech from images is possible. 


In order to evaluate whether this work scales to verbs and adjectives, we compile word lists containing these parts-of-speech in several languages based on a word association dataset (SimLex-999). We collect image sets for each image word and represent all words in a visual space. Then, we rank translations computing similarities between image sets and evaluate performance on this task.

Another field of research that exploits image data for NLP applications is the induction of multi-modal embeddings, i.e. semantic representations that are learned by integrating textual and visual information \cite{Kiela2014,HillKorhonen2014,KielaBottou2014,Lazaridou2015,Silberer2016}. The work presented in our paper differs from these approaches, in that we do not use image data to improve semantic representations, but use images as a resource to learn cross-lingual representations.

\subsection{Contributions}
We introduce a novel dataset for learning cross-lingual word embeddings from images, and we evaluate the approaches by \citet{Bergsma2011Vis} and \newcite{Kiela2015} on this data set, which apart from nouns includes both adjectives and verbs. In addition to considering representations of image words through sets of images, we also experiment with representations that integrate cross-lingual word-embeddings learned from text. Our results, nevertheless, suggest that none of the approaches involving image data are directly applicable to learning cross-lingual representations for adjectives and verbs. In sum, our paper is a {\em negative result}~ paper, showing that what seems like a promising approach to learning bilingual dictionaries for low-resource languages, does not scale beyond simple nouns.

\section{Data}\label{sec:data}
\paragraph{Simlex-999}
We use the Simlex-999 data set of English word pairs \cite{simlex} to compile the word lists for our experiments. The dataset contains words of three part-of-speech classes (nouns, adjectives and verbs) comprising different levels of concreteness. We collect all distinct words in the dataset and randomly sample half of the words for each part-of-speech. The final word list comprises 557 English words (375 nouns, 116 verbs and 66 adjectives). We translate the word list into 5 languages (German, Danish, French, Russian, Italian) using the Yandex translation API\footnote{https://translate.yandex.com/} and manually refine translations.\footnote{The Italian word list is compiled using the translated Simlex-999 data set \cite{Leviant2015}.}

\paragraph{Image Data Sets}
Using the Bing Image Search API\footnote{https://www.microsoft.com/cognitive-services/en-us/bing-image-search-api}, we represent each word in a word list by a set of images by collecting the first 50 jpeg images returned by the search engine when querying the word. This way, we compile image data sets for 6 languages. Identical images contained in image sets for two different languages are removed.\footnote{We completely remove translation pairs for which more than 10 images are identical across languages. Even though identical images returned by the search engine are a strong hint that two words are translations of each other, we want to eliminate the effect of possible cross-lingual indexing internally applied by the search engine.} Figure 1 shows examples for images associated with a word in two languages.





\begin{figure*}[ht]
\resizebox{\textwidth}{!}{
\subfigure[Images associated with the English noun \textit{cow} (left) and the German translation \textit{Kuh} (right).]{\begin{minipage}[b]{.23\textwidth}\includegraphics[width=1\linewidth]{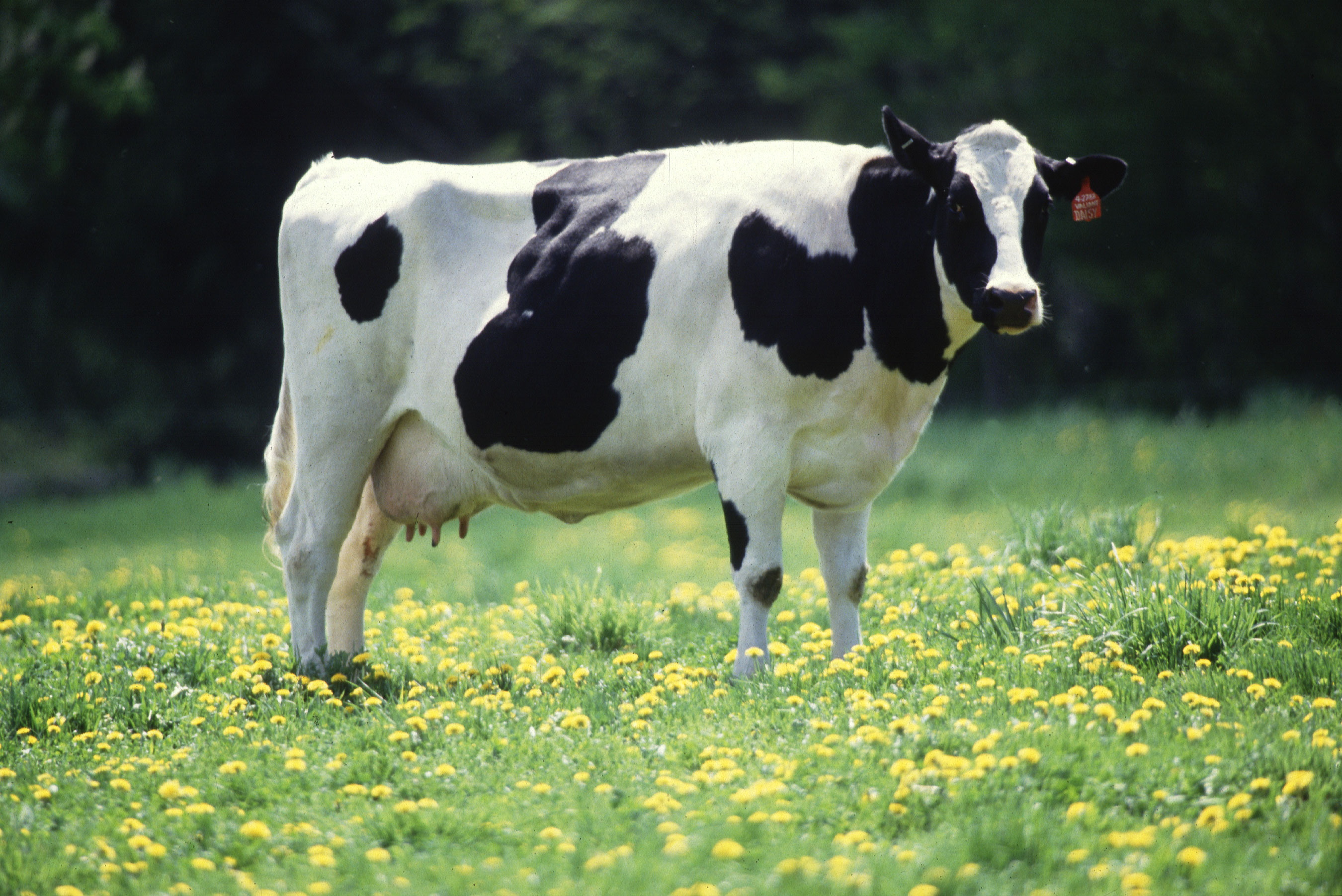}\end{minipage} \quad \begin{minipage}[b]{.23\textwidth}\includegraphics[width=1\linewidth]{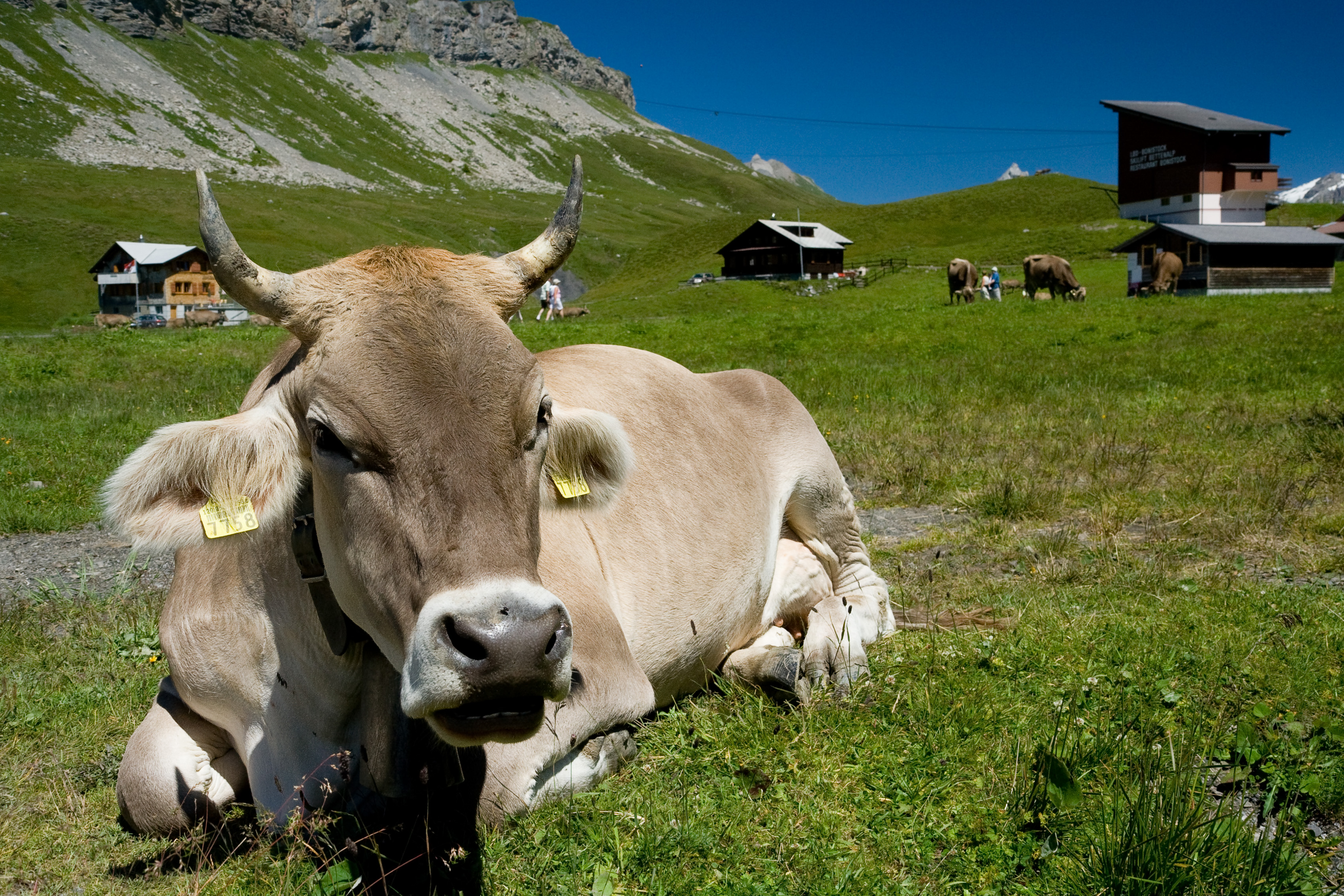}\end{minipage}} \quad \subfigure[Images associated with the English verb \textit{discuss} (left) and the German translation \textit{diskutieren} (right).]{\begin{minipage}[b]{.23\textwidth}\includegraphics[width=1\linewidth]{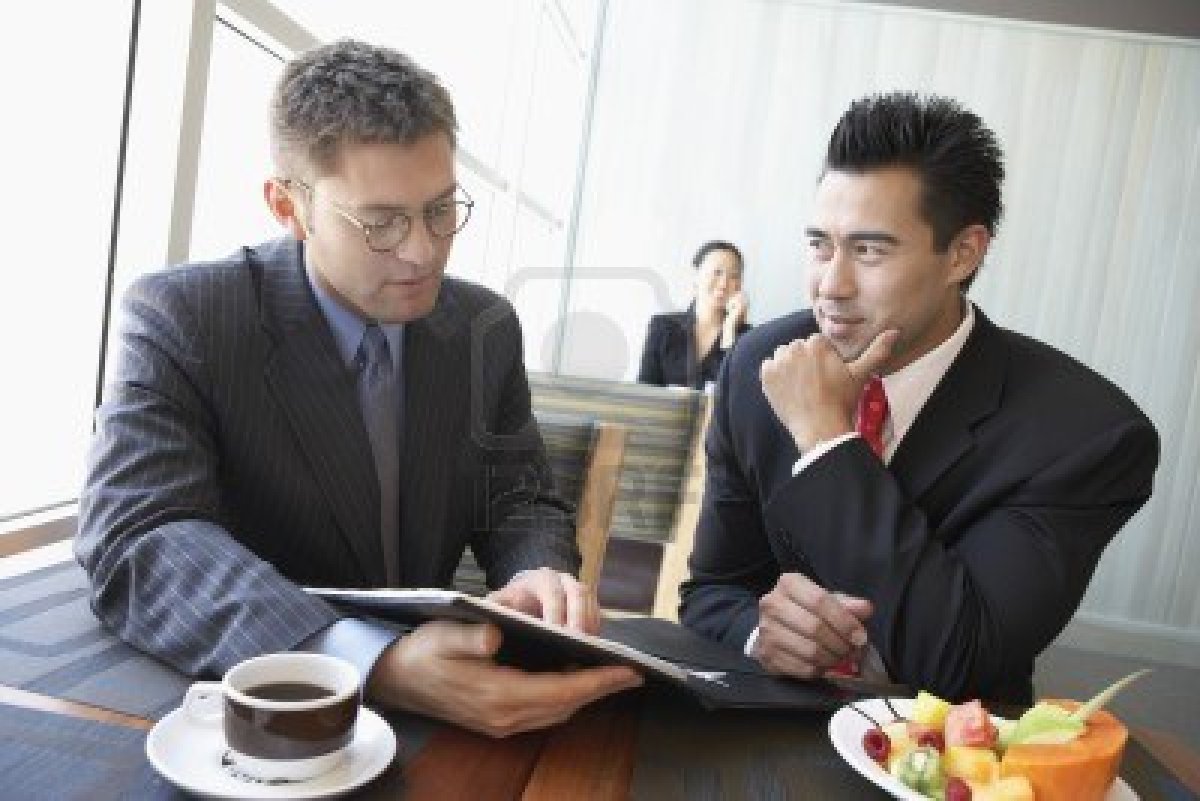}\end{minipage}\quad \begin{minipage}[b]{.23\textwidth}\includegraphics[width=1\linewidth]{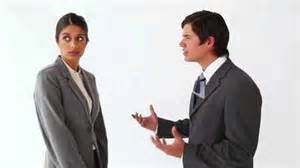}\end{minipage}} \quad \subfigure[Images associated with the English adjective \textit{sad} (left) and the German translation \textit{traurig} (right).]{\begin{minipage}[b]{.23\textwidth}\includegraphics[width=1\linewidth]{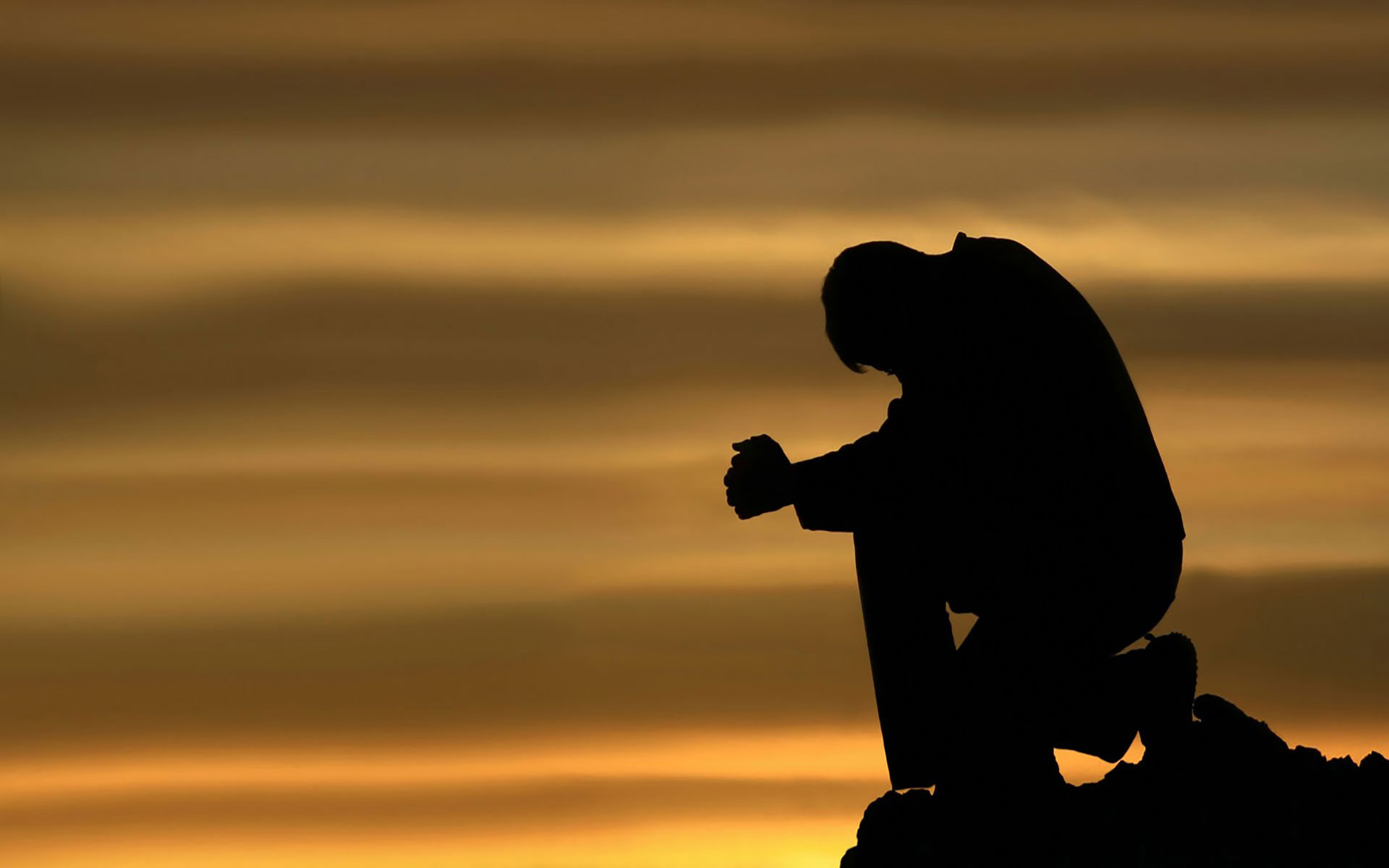}\end{minipage} \quad \begin{minipage}[b]{.23\textwidth}\includegraphics[width=1\linewidth]{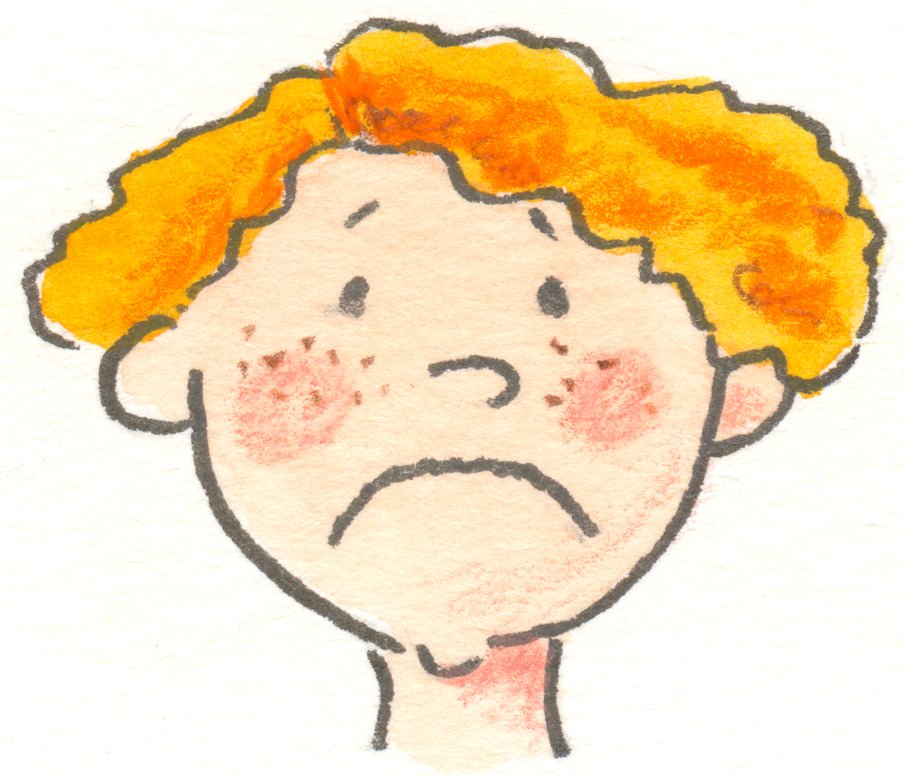}\end{minipage}}
}
\caption{Examples for images associated with equivalent words in two languages (English and German). }
\end{figure*}


\section{Approach}

The assumption underlying the approach is that semantically similar words in two languages are associated with similar images. Hence, in order to find the translation of a word, e.g. from English to German, we compare the images representing the English word with all the images representing German words, and pick as translation the German word represented by the most similar images. To compute similarities between images, we compute cosine similarities between their feature representations.

Following previous work, we obtain two different feature representations for each image, color and SIFT features as proposed by \citet{Bergsma2011Vis} and convolutional network features as proposed by \citet{Kiela2015}. 

\subsection{SIFT and Color Features}
SIFT features \cite{Lowe2004} are descriptors for points of interest in an image. By matching them with descriptor representations in a codebook, images can be represented by a bag-of-codewords from the codebook \cite{Sivic2003}. We apply the optimal hyperparameters reported by Bergsma and Van Durme and generate a codebook by clustering 20,000 codewords from 430,000 randomly sampled descriptors in the English image data. Color features are implemented as color histograms for the red, blue and green channels as well as a gray-scale histogram.

\subsection{Convolutional Neural Network Feature Representations} Following \citet{Kiela2015}, we compute convolutional neural network (CNN) feature representations. We refer to this representation as \visual. For each image, we extract its pre-softmax layer representation in an AlexNet \cite{Krizhevsky2012} pre-trained on the ImageNet classification task. 

\subsection{Representations from Cross-Lingual Word-Embeddings}

In order to investigate if it is possible to learn cross-lingual representations for verbs and adjectives from image data, we also want to know how well text-based methods perform in this task. Hence, we extend our experiments to cross-lingual word-embeddings trained on text data. We use the pre-trained multilingual word-embeddings provided by \citet{Ammar2016}.\footnote{http://128.2.220.95/multilingual/data/} These word-embeddings are generated by first training word-embeddings in a monolingual space. Then, these representations are mapped into a joint multi-lingual space maximizing the similarity between representations that are translations according to a bilingual dictionary. 

Note that since this cross-lingual representation learning algorithm requires a dictionary in advance, it does not scale to the low-resource language scenario that we consider here. We nevertheless include these embeddings, because they strengthen our negative result. The image-based representations do not scale to adjectives and verbs, not even as a supplement to text-based cross-lingual representations. 

For our experiments, we generate a text-based representation for each image by extracting the multi-lingual word-embedding for the image word associated with the image. Note that this results in all images in an image set having the same text-based representation. We refer to this representation as \textual. 

\subsection{Combined Representations}
Besides purely image-based and purely text-based representations, we also consider combined representations that integrate the visual features with the information from text-based word-embeddings. In order to obtain an image's combined representation, we concatenate its CNN feature vector with the text-based representation of its associated image word. This representation is referred to as \combi. 

As the CNN features have dimensionality 4096 and the word-embeddings have dimensionality 40, we also compute a version of the CNN features with reduced dimensionality (40 dimensions) using principal component analysis (referred to as \visualPca). We then generate a combined representation by concatenating the 40-dimensional CNN feature-vector with the 40-dimensional word-embedding, resulting in an 80-dimensional combined representation. This representation is referred to as \combiPca.

\subsection{Similarity Computation}\label{s:similarityComputation}

We implement several methods to determine similarities between representations. 

\paragraph{Similarities Between Individual Images}
\citet{Bergsma2011Vis} determine similarities between image sets based on similarities between all individual images. For each image in image set 1, the maximum similarity score for any image in image set 2 is computed. These maximum similarity scores are then either averaged (\textsc{AvgMax}) or their maximum is taken (\textsc{MaxMax}).

\paragraph{Similarities Between Aggregated Representations}
In addition to the above described methods, \citet{Kiela2015} generate an aggregated representation for each image set and then compute the similarity between image sets by computing the similarity between the aggregated representations. Aggregated representations for image sets are generated by either taking the component-wise average (\textsc{CNN-Mean}) or the component-wise maximum (\textsc{CNN-Max}) of all images in the set.

\paragraph{K-Nearest Neighbor}
For each image in an image set in language 1, we compute its nearest neighbor across all image sets in language 2. Then, we find the image set in language 2 that contains the highest number of nearest neighbors. The image word is translated into the image word that is associated with that image 2 set. Ties between image sets containing an equivalent number of nearest neighbors are broken by computing the average distance between all members. We refer to the method as \textsc{KNN}.

Whereas the other approaches described above provide a ranking of translations, this method determines only the one translation that is associated with the most similar image set.

\paragraph{Clustering Image Sets}
As we expect the retrieved image sets for a word to contain images associated with different senses of the word, we first cluster images into $k$ clusters. This way, we hope to group images representing different word senses. Then, we apply the \textsc{KNN} method as described above. We refer to this method as \textsc{KNN-C}.

\paragraph{Logistic Regression}
The above described methods are unsupervised and do not require parameter estimation. In order to check if performance can be increased using supervision, we train a logistic regression ranking model (\textsc{LogRegr}) to predict rankings of translations. Training data is generated by pairing all image sets across two languages and computing difference vectors between the aggregated mean representations of each image set. Pairs of image sets that correspond to correct translation pairs are labeled as positive while all other pairs are labeled as negative. In the prediction step, translations are ranked based on the predicted distance to the hyperplane \cite{Joachims2002c}.

\subsection{Evaluation Metrics}
Ranking performance is evaluated by computing the Mean Reciprocal Rank (MRR) as
\[MRR = \frac{1}{M}\sum\limits^M_{i=1}\dfrac{1}{rank(w_s, w_t)}\]
$M$ is the number of words to be translated and $rank(w_s, w_t)$ is the position the correct translation $w_t$ for source word $w_s$ is ranked on. 

In addition to MRR, we also evaluate the cross-lingual representations by means of precision at $k$ (P@$k$). For the nearest neighbor approach, we can only compute P@1.

\begin{table*}[ht]
\centering
\resizebox{\textwidth}{!}{
\begin{tabular}{l|rrr|rrr|rrr|rrr}
\toprule[1pt]
 & \multicolumn{3}{c}{\textbf{ALL}} & \multicolumn{3}{|c|}{\textbf{NN}} & \multicolumn{3}{|c|}{\textbf{VB}} & \multicolumn{3}{|c}{\textbf{ADJ}}\\
\cmidrule{2-13}
 & MRR & P@1 & P@10 & MRR & P@1 & P@10 & MRR & P@1 & P@10 & MRR & P@1 & P@10\\
\cmidrule{1-13}
\textsc{avgMax}  & 0.40 &0.34 &0.50 & 0.56 &0.49 &0.67 &
 0.06 &0.02 &0.12 & 0.15 &0.10 &0.25 \\
\textsc{maxMax}  & 0.31 &0.24 &0.42 & 0.43 &0.34 &0.58 & 0.05 &0.03 &0.07 & 0.10 &0.06 &0.16 \\
\textsc{CnnMean}  &0.35 &0.29 &0.45 & 0.48 &0.41 &0.60 & 0.07 &0.02 &0.13 & 0.14 &0.09 &0.22 \\
\textsc{CnnMax}  & 0.33 &0.27 &0.44 & 0.47 &0.39 &0.60 & 0.06 &0.01 &0.11 & 0.11 &0.07 &0.20 \\
\textsc{KNN}& -- & 0.28 & -- & -- & 0.40 & -- & --& 0.01 & -- & -- & 0.06 & --\\
\textsc{KNN-C} ($k=3$)& -- & 0.28 & -- & -- & 0.40 & -- & --& 0.01 & -- & -- & 0.04 & --\\
\cmidrule{1-13}
\textsc{LogRegr}& 0.23 & 0.21 & 0.35 & 0.37 & 0.31 & 0.48 & 0.04 & 0.01 & 0.08 & 0.08 & 0.04 & 0.12\\
\addlinespace[5pt]
\end{tabular}}
\caption{Results for translation ranking with images represented by CNN features averaged over 5 language pairs. \textsc{KNN} and \textsc{KNN-C} do not produce a ranking, hence we only provide P@1 values. Results for \textsc{LogRegr} are averaged over two splits, i.e. each split used for testing contains only half of the words. Hence the MRR values are not comparable to the other settings listed in the table.}\label{t:cnnFeatures}
\end{table*}

\begin{table*}[h]
\centering
\resizebox{\textwidth}{!}{
\begin{tabular}{l|rrr|rrr|rrr|rrr}
\toprule[1pt]
 & \multicolumn{3}{|c|}{\textbf{ALL}} & \multicolumn{3}{|c|}{\textbf{NN}} & \multicolumn{3}{|c|}{\textbf{VB}} & \multicolumn{3}{|c}{\textbf{ADJ}}\\
\cmidrule{2-13}
 & MRR & P@1 & P@10 & MRR & P@1 & P@10 & MRR & P@1 & P@10 & MRR & P@1 & P@10\\
\cmidrule{1-13}
\textsc{avgMax}  & 0.06 &0.04 &0.10 &0.08 &0.05 &0.13 &
 0.03 &0.02 &0.04 & 0.01 &0.00 &0.01 \\
\textsc{maxMax}  & 0.06 &0.04 &0.08 & 0.07 &0.04 &0.11 & 0.04 &0.02 &0.04 & 0.01 &0.00 &0.03 \\
\textsc{CnnMean}  &0.07 &0.04 &0.13 & 0.10 &0.06 &0.18 & 0.01 &0.00 &0.01 & 0.02 &0.00 &0.04 \\
\textsc{CnnMax}  & 0.04 &0.02 &0.07 & 0.06 &0.03 &0.10 & 0.01 &0.00 &0.02 & 0.02 &0.00 &0.01 \\
\addlinespace[5pt]

\end{tabular}}
\caption{Results for translation ranking with images represented by SIFT and color features averaged over 5 language pairs.}\label{t:siftFeatures}
\end{table*}

\begin{figure*}[h]
\centering
\resizebox{\textwidth}{!}{
\subfigure{\begin{minipage}[b]{.23\textwidth}\includegraphics[width=1\linewidth]{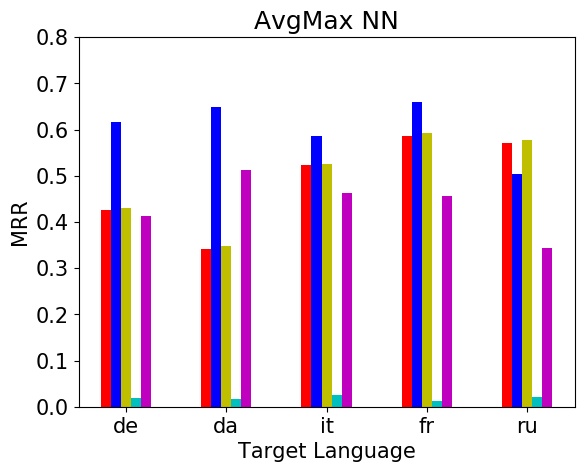}\end{minipage} \quad \begin{minipage}[b]{.23\textwidth}\includegraphics[width=1\linewidth]{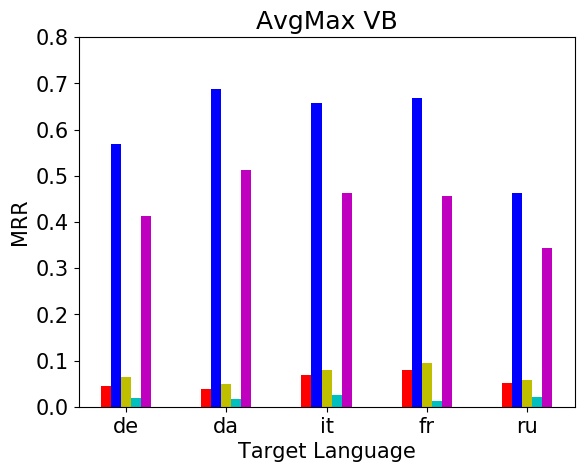}\end{minipage}  \quad \begin{minipage}[b]{.23\textwidth}\includegraphics[width=1.14\linewidth]{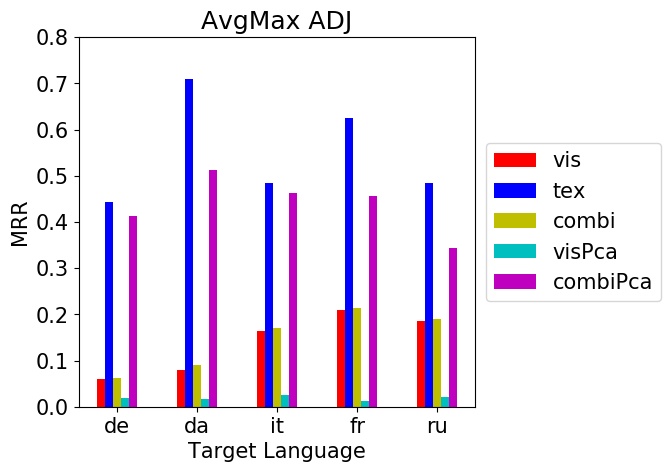}\end{minipage}} 
}
\caption{MRR results for the \textsc{avgMax} similarity computation on the combined representations (\combi) and the representations after applying PCA (\visualPca~and \combiPca) compared to the purely visual and textual representations (\visual~and \textual) for each part-of-speech and each language pair.}\label{f:CombiAvgMax}
\end{figure*}

\begin{figure*}[h]
\centering
\resizebox{\textwidth}{!}{
\subfigure{\begin{minipage}[b]{.23\textwidth}\includegraphics[width=1\linewidth]{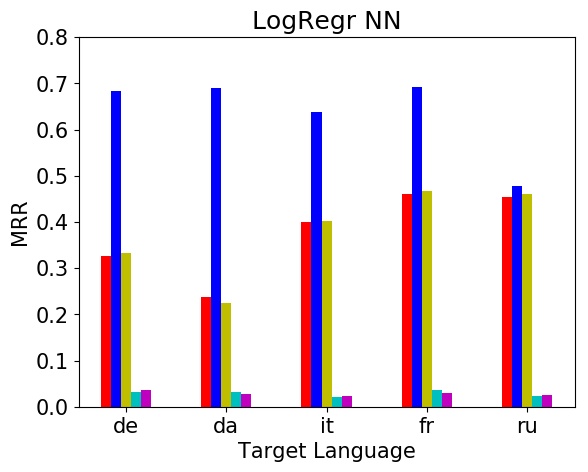}\end{minipage} \quad \begin{minipage}[b]{.23\textwidth}\includegraphics[width=1\linewidth]{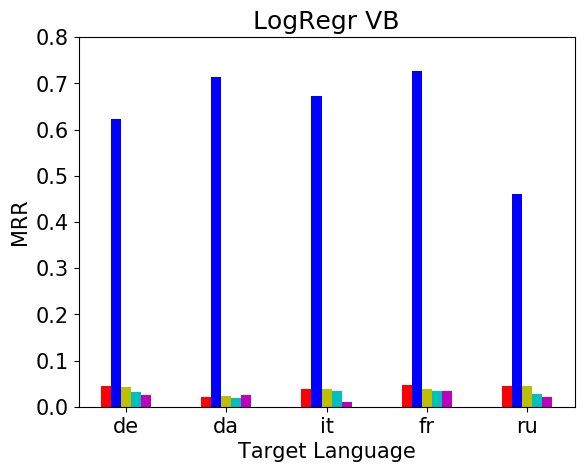}\end{minipage}  \quad \begin{minipage}[b]{.23\textwidth}\includegraphics[width=1.14\linewidth]{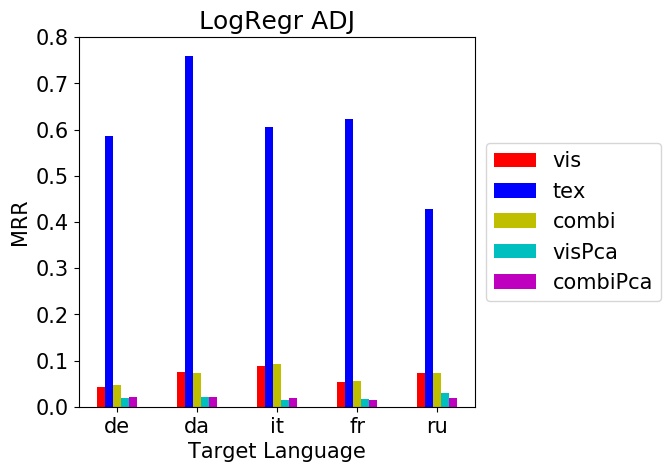}\end{minipage}} 
}
\caption{MRR results for the translation ranking using the \textsc{LogRegr} model on the combined representations (\combi) and the representations after applying PCA (\visualPca~and \combiPca) compared to the purely visual and textual representations (\visual~and \textual) for each part-of-speech and each language pair.}\label{f:CombiLogReg}
\end{figure*}

\section{Experiments and Results}
We run experiments for 5 language pairs English--German, English--Danish, English--French, English--Russian and English--Italian. In the first experiment, we consider only the representations computed from image data (SIFT + color and CNN features) and compare the different methods for similarity computation described in Section \ref{s:similarityComputation}. For each English word, we rank all the words in the corresponding target languages based on similarities between image sets and evaluate the models' ability to identify correct translations, i.e. to rank the correct translation on a position near the top. We compare 4 settings that differ in the set of English words that are translated. In the setting \textsc{ALL}, all English words in the word list are translated. \textsc{NN}, \textsc{VB} and \textsc{ADJ} refer to the settings where only nouns, verbs and adjectives are translated. 

In a second experiment, we apply the best performing similarity computation method on the visual representations (\textsc{avgMax}) and the \textsc{LogRegr} model to the visual, textual and combined representations.

\subsection{Results}

We present results for the comparison between different similarity computation methods for purely visual representations in Tables \ref{t:cnnFeatures} and \ref{t:siftFeatures}. Figures \ref{f:CombiAvgMax} and \ref{f:CombiLogReg} show the results for the performance of \textsc{avgMax} and \textsc{LogRegr} on different representations.

\paragraph{Comparison of Similarity Computation Methods on Visual Representations}

Table \ref{t:cnnFeatures} displays results for images represented by CNN features averaged over all language pairs. Table \ref{t:siftFeatures} shows results for images represented by SIFT and color features averaged over all language pairs. 

First, we observe that our results for translating nouns using CNN features are slightly lower than those reported in previous work (\citet{Kiela2015} report P@1 = 0.567 for a dataset containing 500 nouns). This may be due to a higher proportion of more abstract nouns in our dataset. Second, our experiments confirm that the CNN features outperform the color and SIFT features by a large margin, which is in line with the findings by Kiela et al.\footnote{Our results using SIFT and color features are very low, even when considering only nouns. This might be due to the fact that we did not tune any parameters but adopted the optimal parameters reported in previous work, while working with a dataset that contains more variance in the images.} 

Most importantly, we witness a very significant drop in performance when moving from nouns to verbs and adjectives. For verbs, we almost never pick the right translation based on the image-based word representations. This behavior applies across all methods for similarity computation. For the CNN features, \textsc{avgMax} outperforms the other methods in almost all cases.

\paragraph{Comparison of Representations}
Figures \ref{f:CombiAvgMax} and \ref{f:CombiLogReg} compare the performance of the \textsc{avgMax} and \textsc{logRegr} methods applied to the combined representations (\combi) and the representations after applying PCA (\visualPca~and \combiPca), as well as the purely visual and textual representations (\visual~and \textual).

Most importantly, we see that the models involving images (\visual, \combi~and the corresponding \textsc{Pca} versions) are always outperformed by the model using representations learned from text data (except for English-Russian noun pairs). This observation applies to both the \textsc{avgMax} method and the \textsc{logRegr} model. As in Tables \ref{t:cnnFeatures} and \ref{t:siftFeatures}, we can see a significant drop in performance for the representations computed on visual data comparing nouns with adjectives and verbs. In contrast, the quality of the rankings based on text-based representations stays approximately the same for different parts-of-speech. 

\subsection{Analysis}
If we try to learn translations from images, integrating verbs and adjectives into the dataset worsens results compared to a dataset that contains only nouns. One possible explanation is that images associated with verbs and adjectives are less suited to represent the meaning of a concept than images associated with nouns. 

\citet{Kiela2015} suppose that lexicon induction via image similarity performs worse for datasets containing words that are more abstract. In order to approximate the degree of abstractness of a concept, they compute the \textit{image dispersion} $d$ for a word $w$ as the average cosine distance between all image pairs in the image set $\{ i_j, \dots, i_n \}$ associated with word $w$ according to
\[d(w) = \dfrac{2}{n(n-1)}\sum\limits_{k<j\leq n} 1- \dfrac{i_j \cdot i_k}{\lvert i_j\rvert \lvert i_k\rvert}\]

In their analysis, \citet{Kiela2015} find that their model performs worse on datasets with a higher average image dispersion. 

Computing the average image dispersion for our data across languages shows that image sets associated with verbs and adjectives have a higher average image dispersion than image sets associated with nouns (nouns: $d = 0.75$, verbs: $d = 0.83$, adjectives: $d = 0.82$).



\begin{table}[h]
\centering
\begin{tabular}{rlrrlr}
\toprule[1pt]
& \multicolumn{2}{c}{Lowest dispersion} & &\multicolumn{2}{c}{Highest dispersion}\\
 \cmidrule{2-6}
  & Word & $d$ & & Word & $d$ \\
\cmidrule{1-6}
 \multirow{3}{*}{\textbf{NN}} & bicycle &0.29 & & second & 0.88\\
 & letter & 0.30 & & condition & 0.86\\
 & racket & 0.37 & & south & 0.86\\
 \cmidrule{1-6}
 \multirow{3}{*}{\textbf{VB}} & send & 0.67 && create & 0.88 \\
 & do & 0.78& & expand & 0.88\\
 & complain & 0.78 & & try &0.87\\
 \cmidrule{1-6}
 \multirow{3}{*}{\textbf{ADJ}} & bold & 0.49&& necessary & 0.87 \\
 & fragile & 0.49 &&  old & 0.87\\
 & modest & 0.57&& huge & 0.87\\ 
\addlinespace[5pt]
\end{tabular}
\caption{English image words associated with the image sets with highest and lowest dispersion scores $d$.}\label{t:exampleWords}
\end{table}

To gain insights into the relation between image dispersion, part-of-speech and the models ability to find correct translation pairs, we manually analyze the image sets with highest and lowest dispersion values in our dataset. Table \ref{t:exampleWords} shows the image words associated with the image sets that have the highest and lowest dispersion values in our English image data.


\begin{figure*}[h]
\centering
\resizebox{\textwidth}{!}{
\subfigure{
\begin{minipage}[b]{.23\textwidth}\includegraphics[width=1\linewidth]{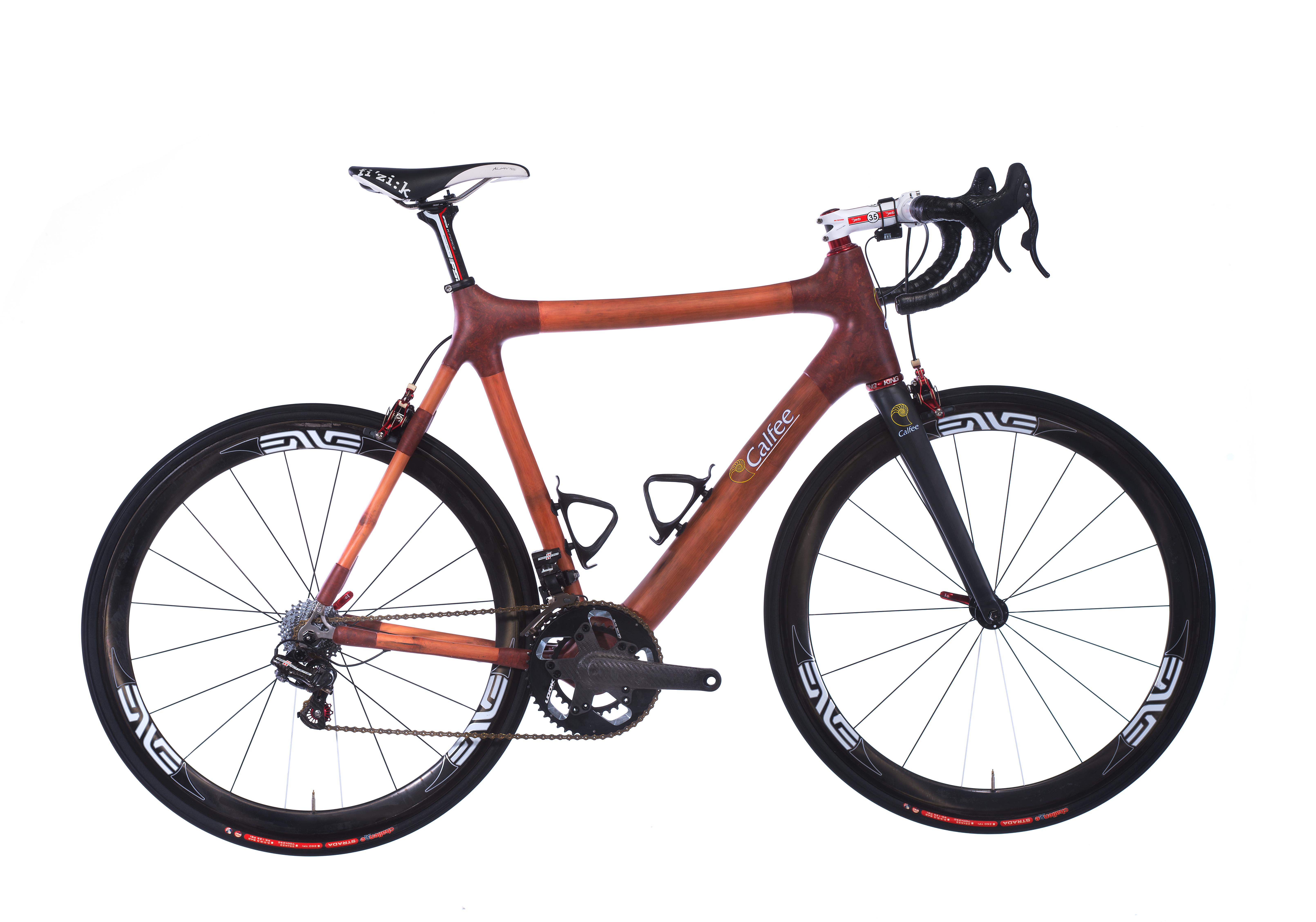}\end{minipage} \quad \begin{minipage}[b]{.23\textwidth}\includegraphics[width=1\linewidth]{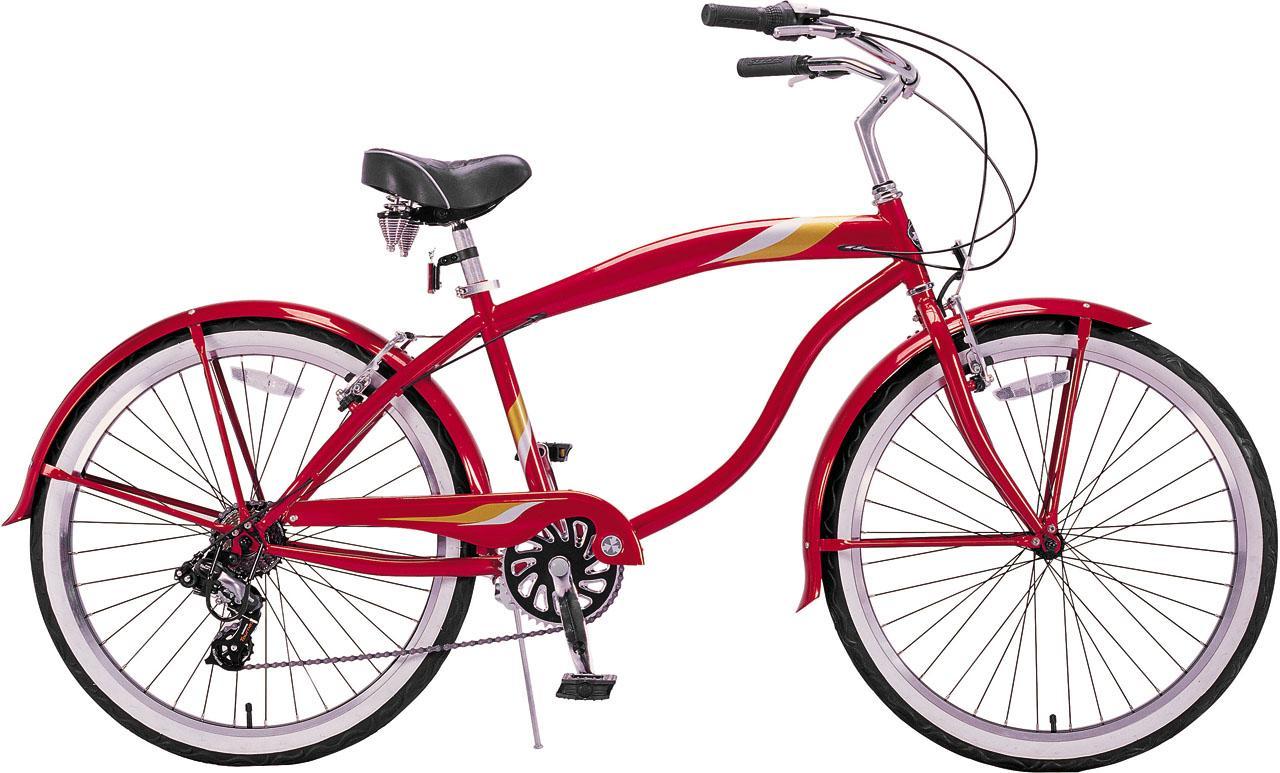}\end{minipage}  \quad \begin{minipage}[b]{.23\textwidth}\includegraphics[width=1\linewidth]{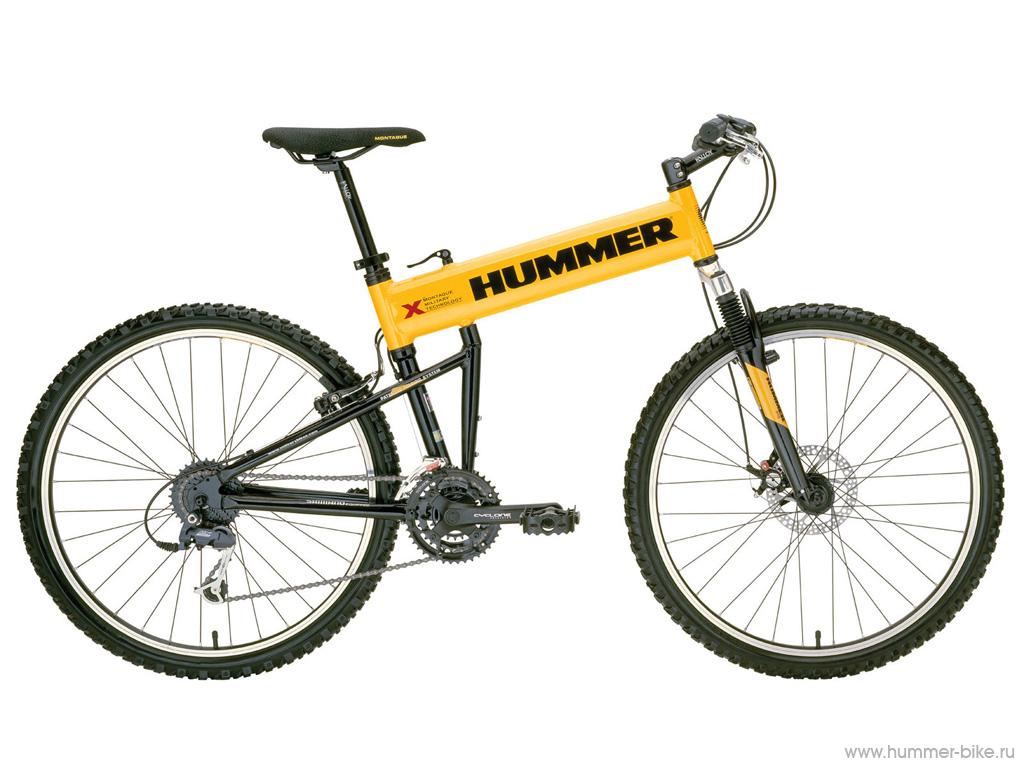}\end{minipage}\quad \begin{minipage}[b]{.23\textwidth}\includegraphics[width=1\linewidth]{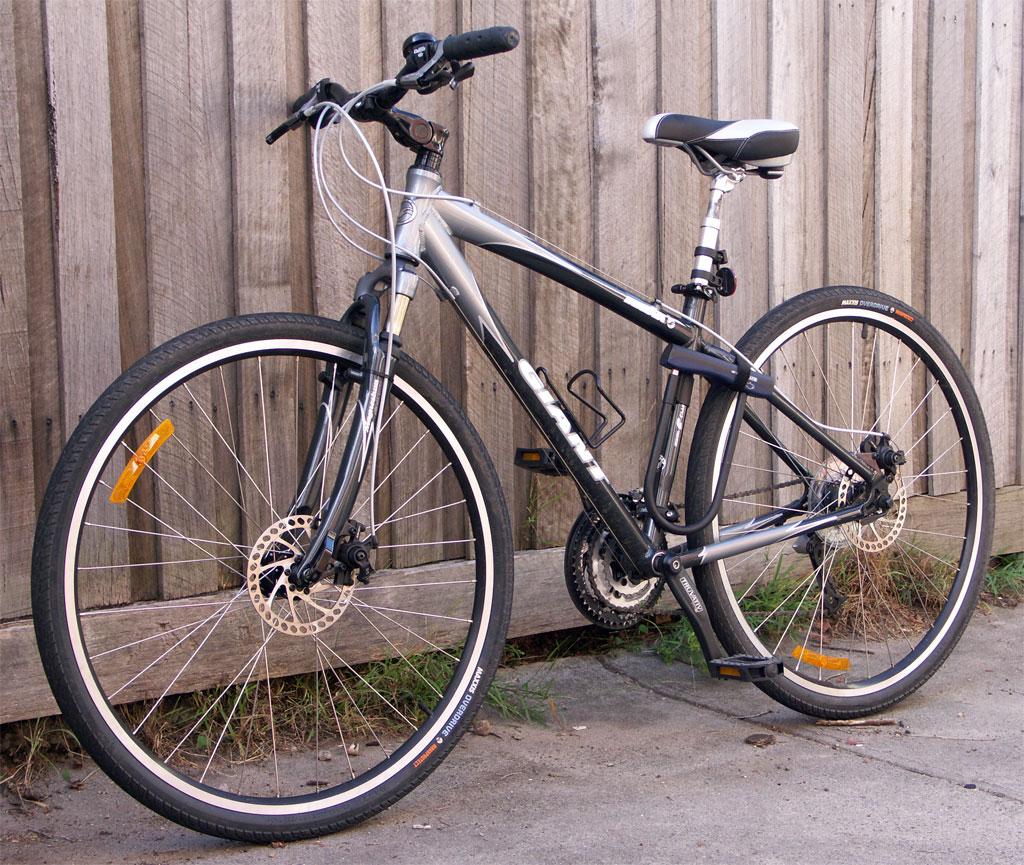}\end{minipage}\quad \qquad \qquad \begin{minipage}[b]{.23\textwidth}\includegraphics[width=1\linewidth]{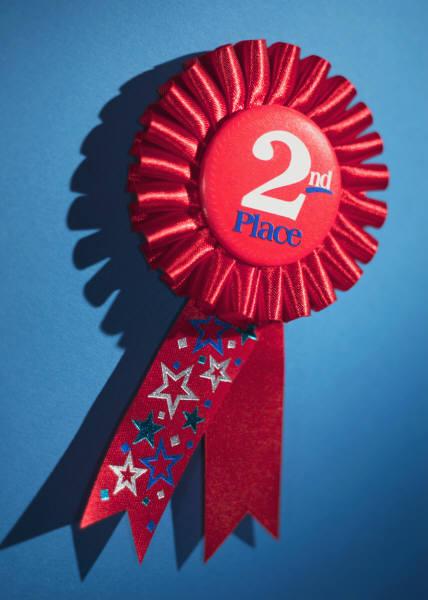}\end{minipage} \quad \begin{minipage}[b]{.23\textwidth}\includegraphics[width=1\linewidth]{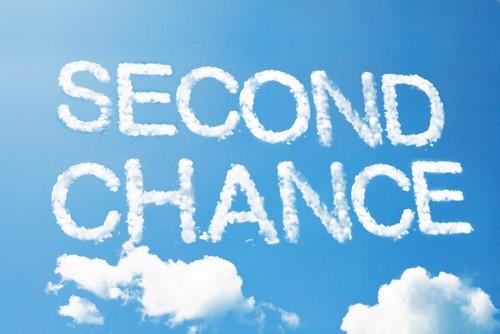}\end{minipage}  \quad  \begin{minipage}[b]{.23\textwidth}\includegraphics[width=1\linewidth]{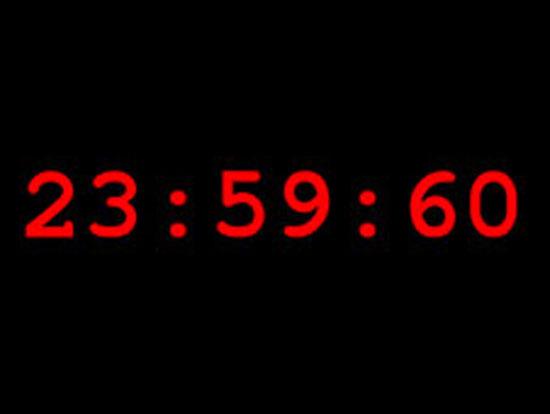}\end{minipage}\quad \begin{minipage}[b]{.23\textwidth}\includegraphics[width=1\linewidth]{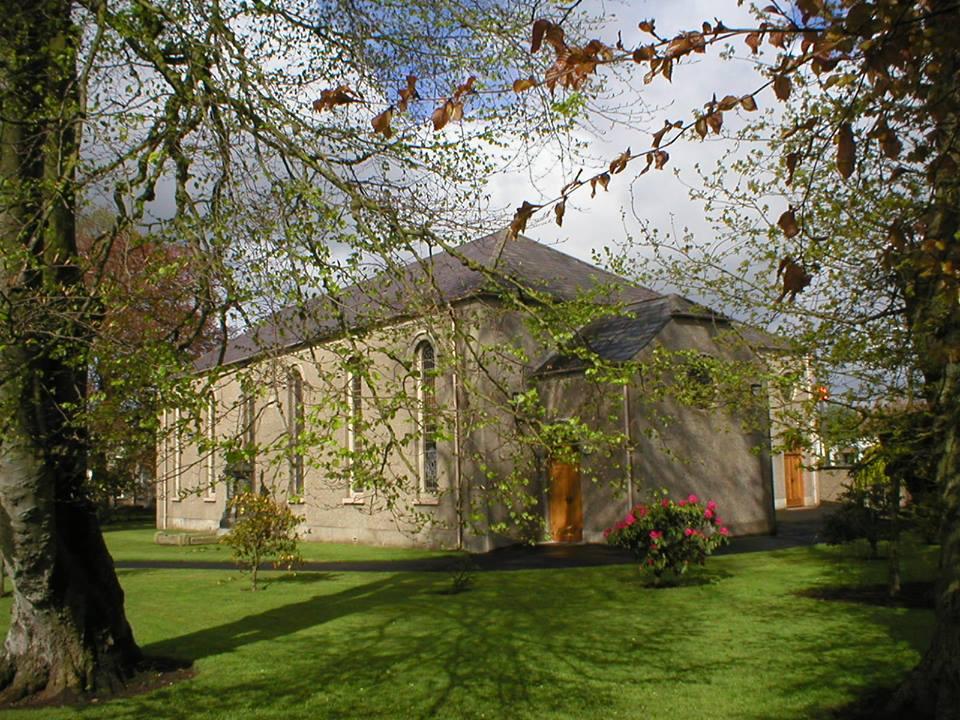}\end{minipage}} 
}
\resizebox{\textwidth}{!}{\subfigure{
\begin{minipage}[b]{.23\textwidth}\includegraphics[width=1\linewidth]{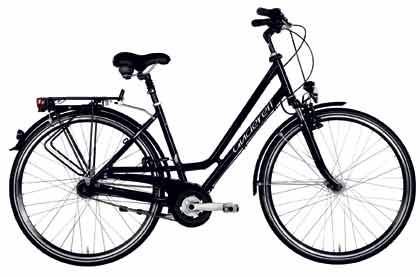}\end{minipage} \quad \begin{minipage}[b]{.23\textwidth}\includegraphics[width=1\linewidth]{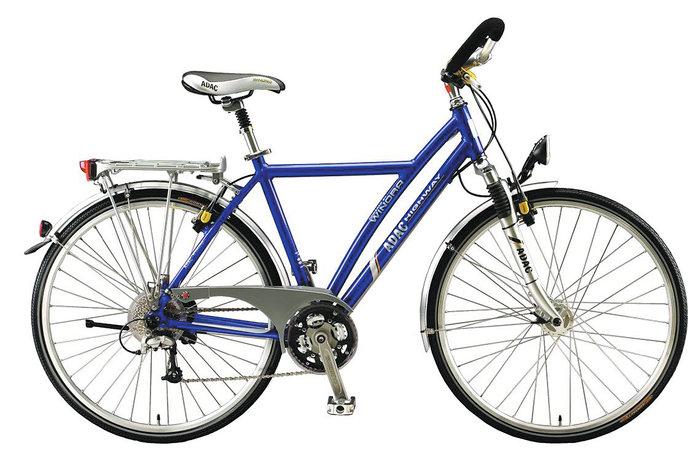}\end{minipage}  \quad \begin{minipage}[b]{.23\textwidth}\includegraphics[width=1\linewidth]{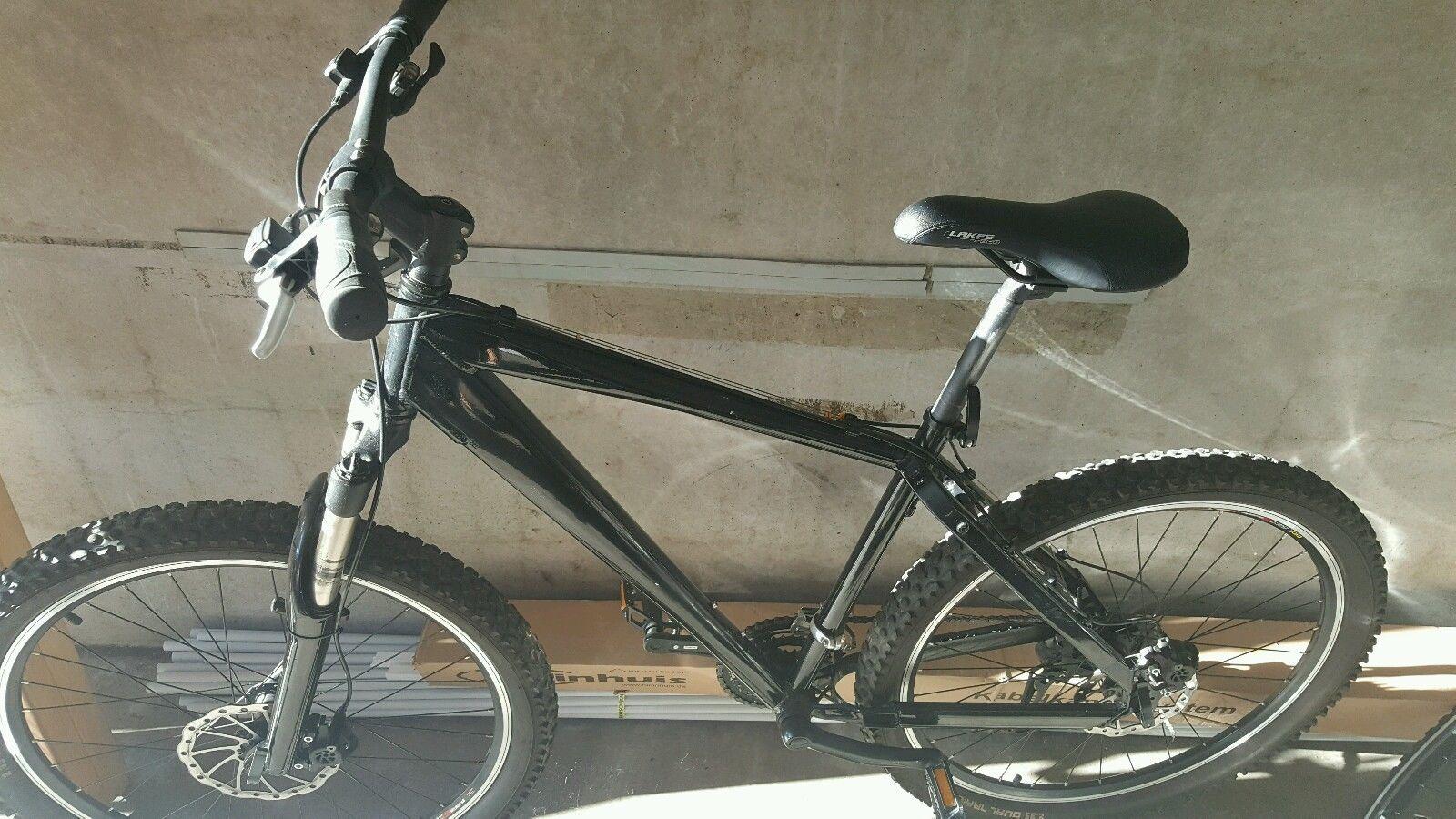}\end{minipage}\quad \begin{minipage}[b]{.23\textwidth}\includegraphics[width=1\linewidth]{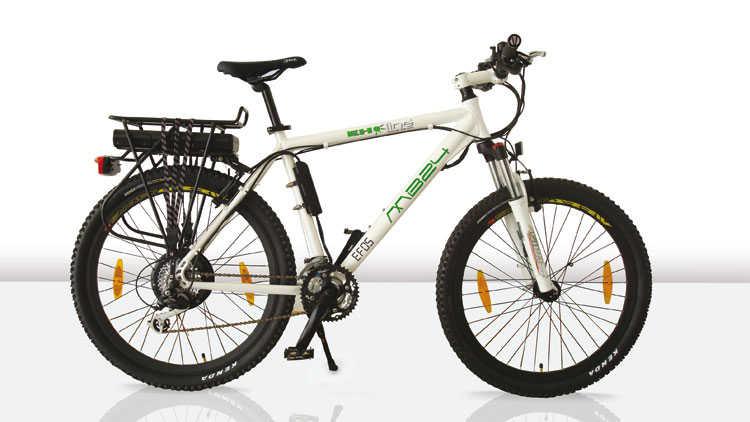}\end{minipage}\quad \qquad \qquad \begin{minipage}[b]{.23\textwidth}\includegraphics[width=1\linewidth]{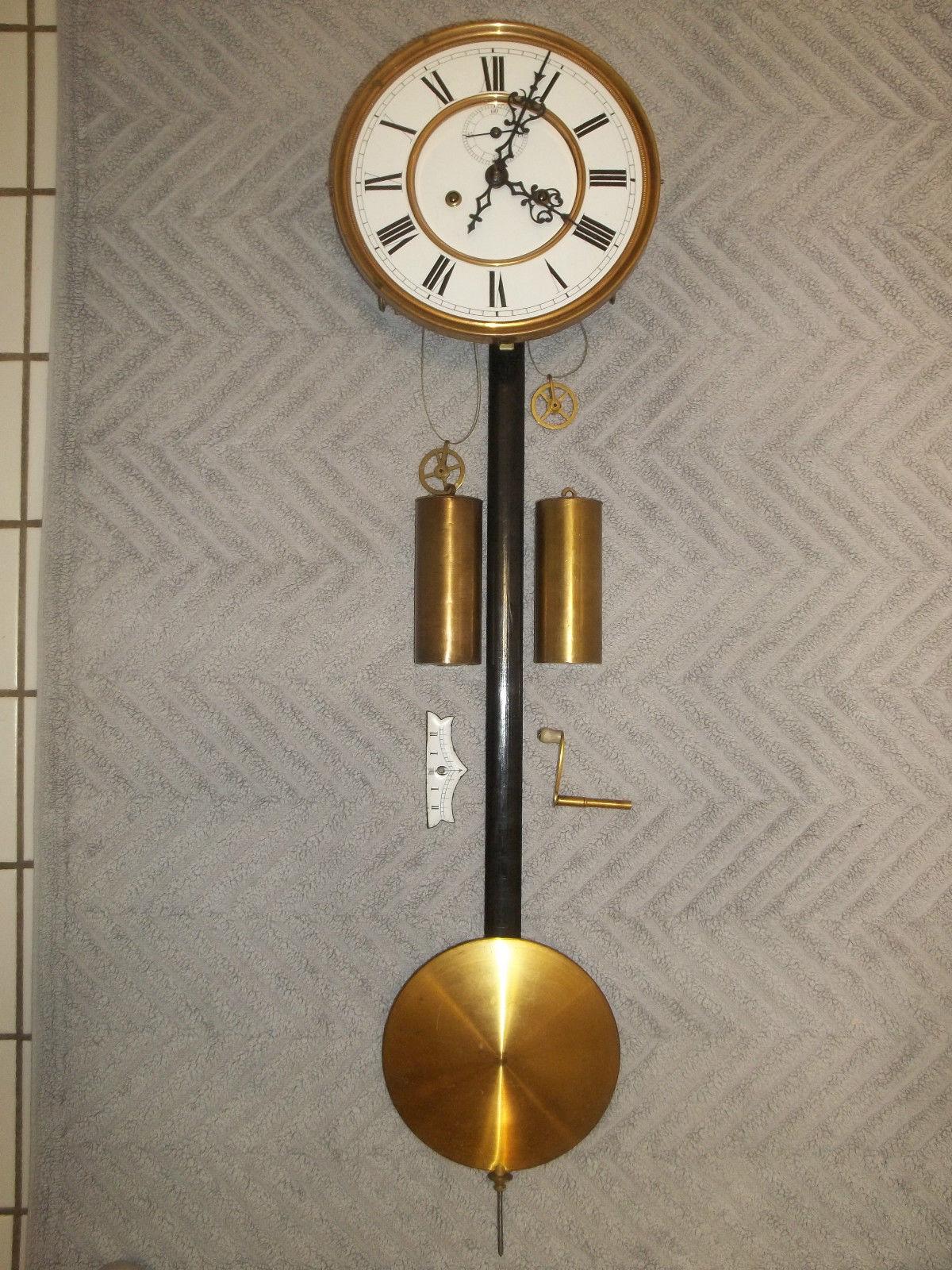}\end{minipage} \quad \begin{minipage}[b]{.23\textwidth}\includegraphics[width=1\linewidth]{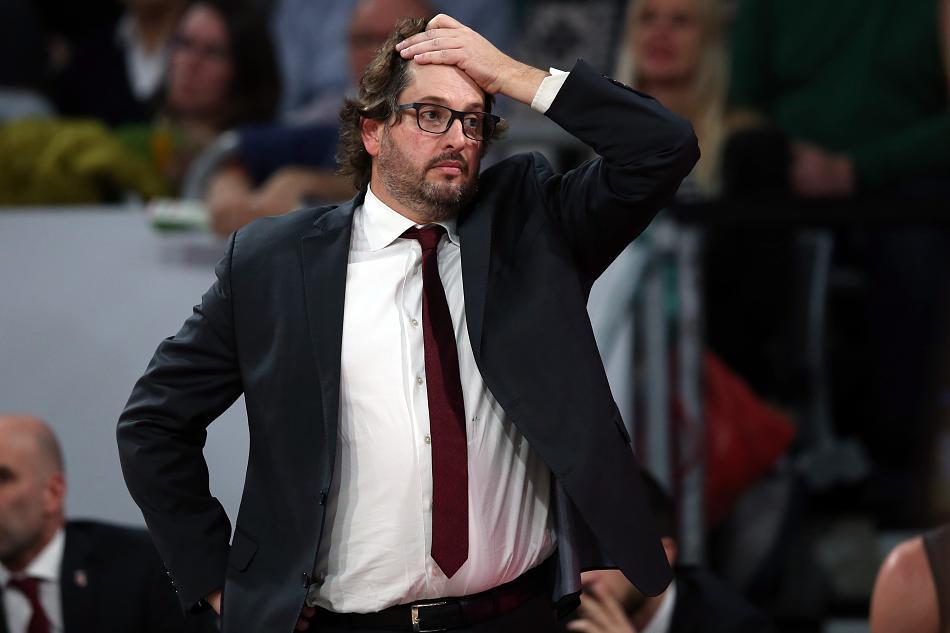}\end{minipage}\quad \begin{minipage}[b]{.23\textwidth}\includegraphics[width=1\linewidth]{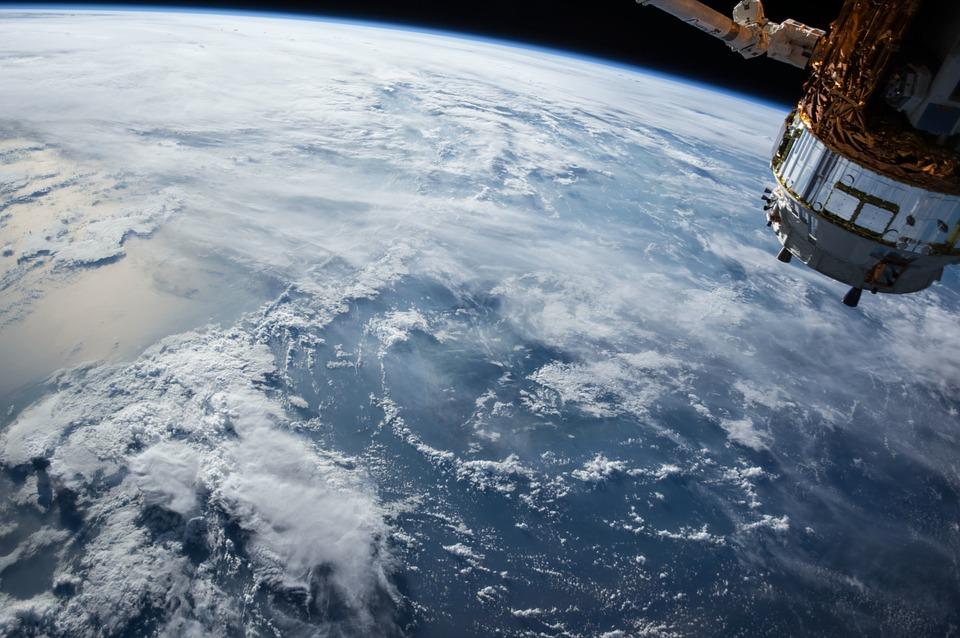}\end{minipage}\quad \begin{minipage}[b]{.23\textwidth}\includegraphics[width=1\linewidth]{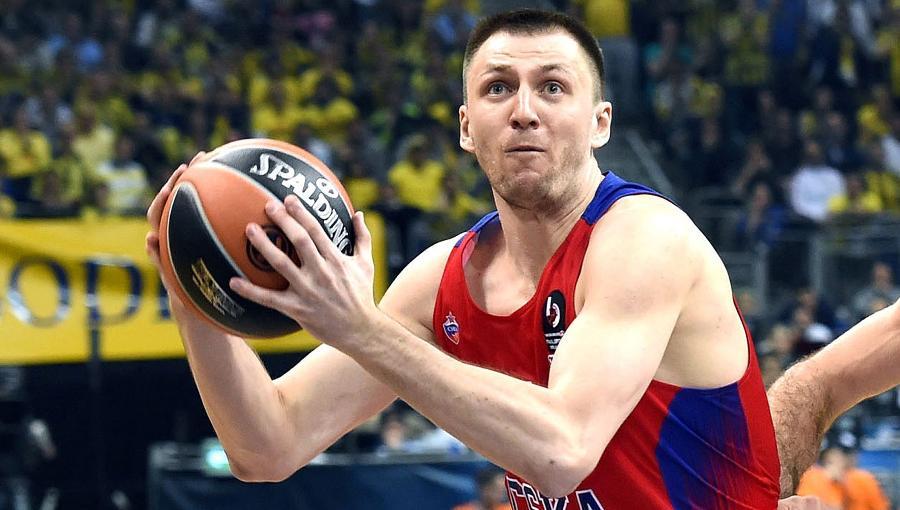}\end{minipage}} 
}
\caption{The first 4 images in the noun image sets with lowest (left) and highest (right) dispersion values, which correspond to \textit{bicycle} and \textit{second} (first row), and the German translations of the English image words, \textit{Fahrrad} and \textit{Sekunde} (second row).}\label{f:imageSetsNouns}
\end{figure*}

\begin{figure*}[h]
\centering
\resizebox{\textwidth}{!}{
\subfigure{
\begin{minipage}[b]{.23\textwidth}\includegraphics[width=1\linewidth]{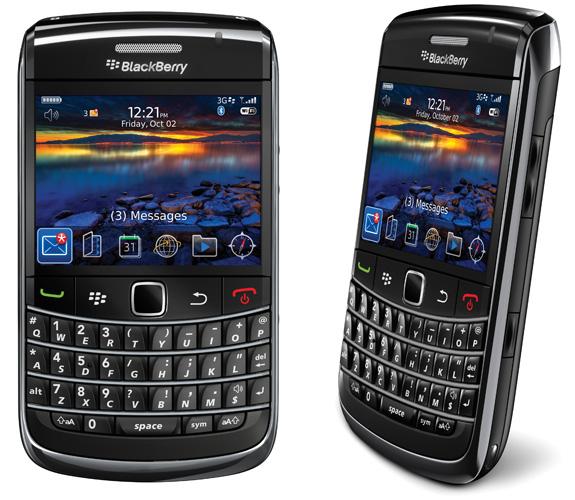}\end{minipage} \quad \begin{minipage}[b]{.23\textwidth}\includegraphics[width=1\linewidth]{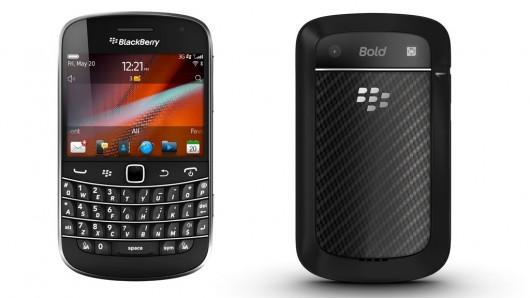}\end{minipage}  \quad \begin{minipage}[b]{.23\textwidth}\includegraphics[width=1\linewidth]{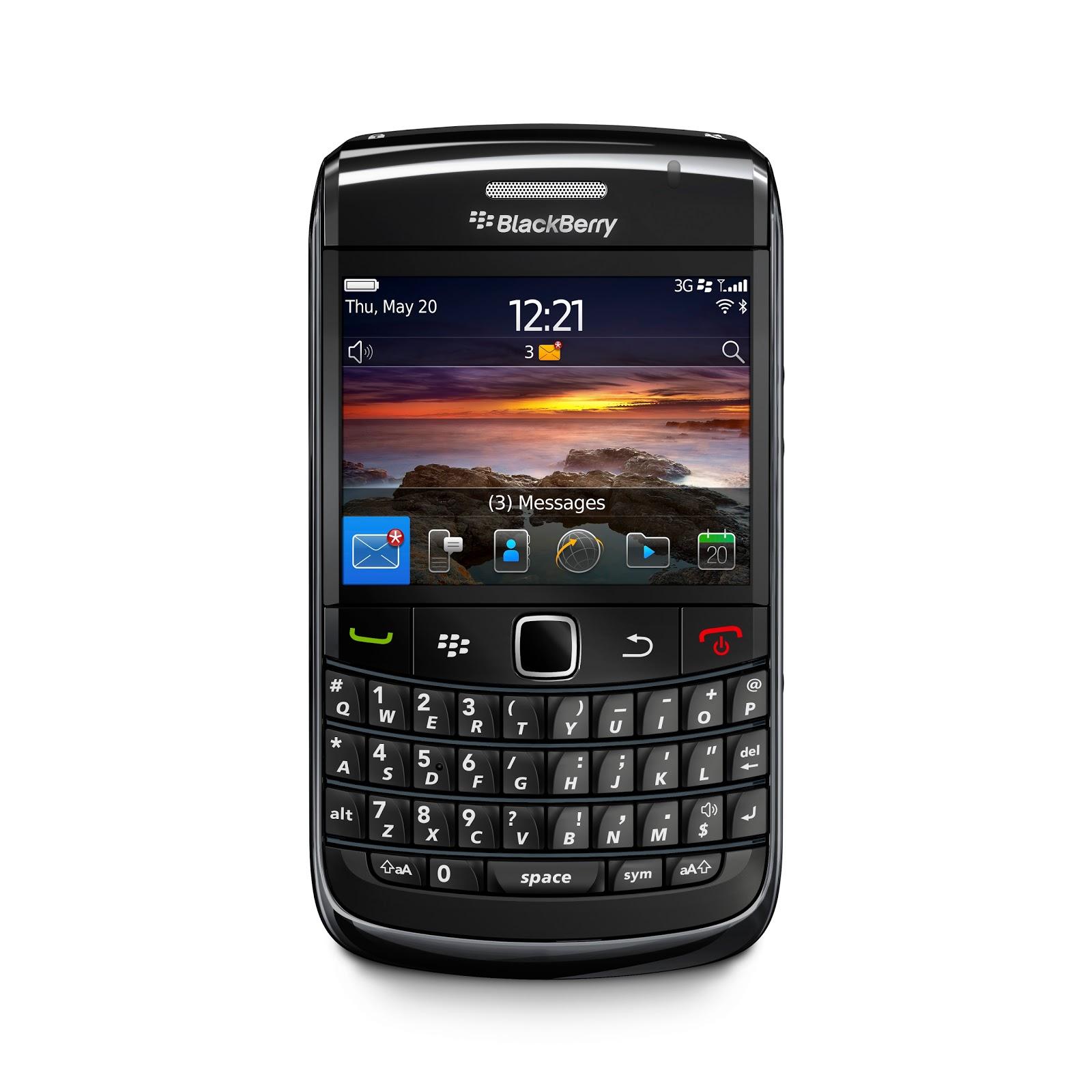}\end{minipage}\quad \begin{minipage}[b]{.23\textwidth}\includegraphics[width=1\linewidth]{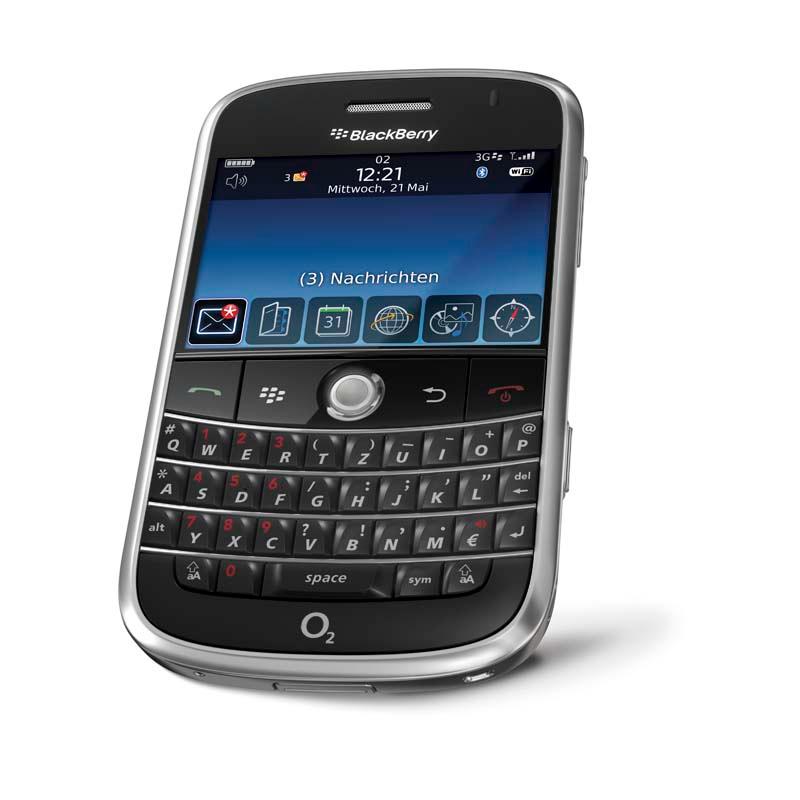}\end{minipage}\quad \qquad \qquad  \begin{minipage}[b]{.23\textwidth}\includegraphics[width=1\linewidth]{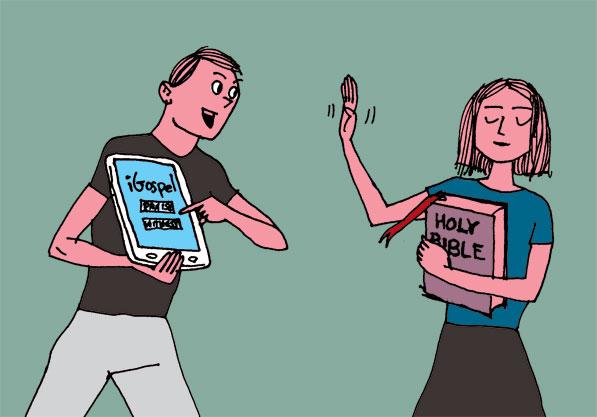}\end{minipage} \quad \begin{minipage}[b]{.23\textwidth}\includegraphics[width=1\linewidth]{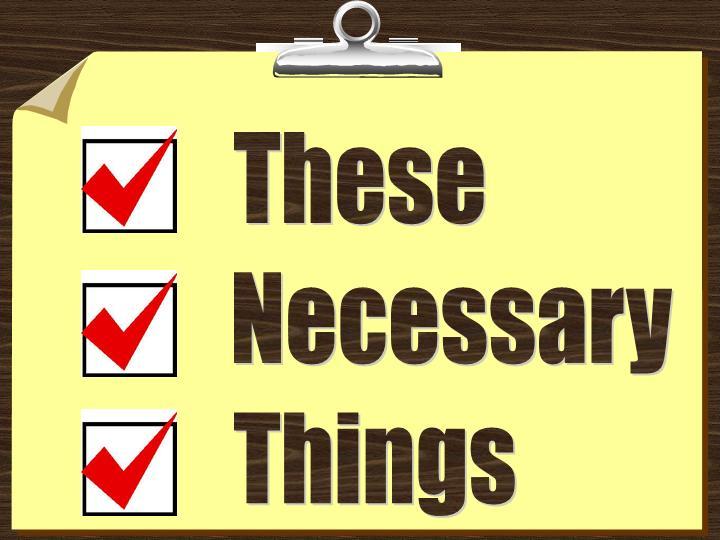}\end{minipage}  \quad \begin{minipage}[b]{.23\textwidth}\includegraphics[width=1\linewidth]{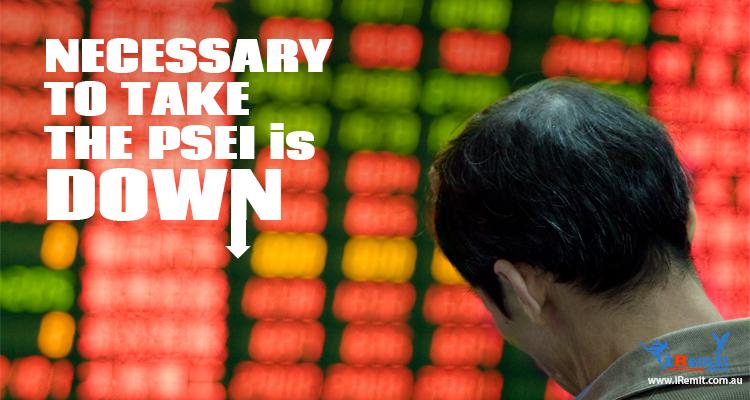}\end{minipage}\quad \begin{minipage}[b]{.23\textwidth}\includegraphics[width=1\linewidth]{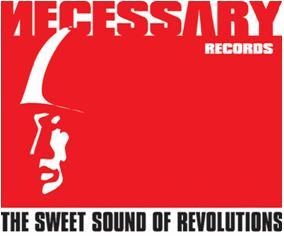}\end{minipage}} 
}
\resizebox{\textwidth}{!}{\subfigure{
\begin{minipage}[b]{.23\textwidth}\includegraphics[width=1\linewidth]{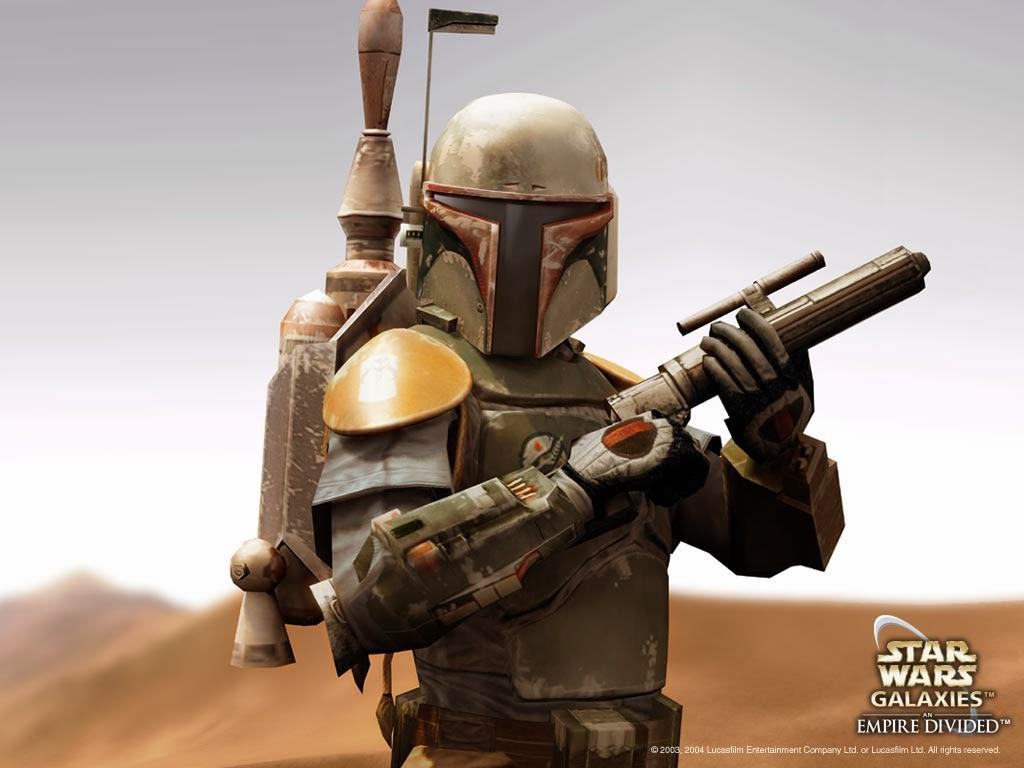}\end{minipage} \quad \begin{minipage}[b]{.23\textwidth}\includegraphics[width=1\linewidth]{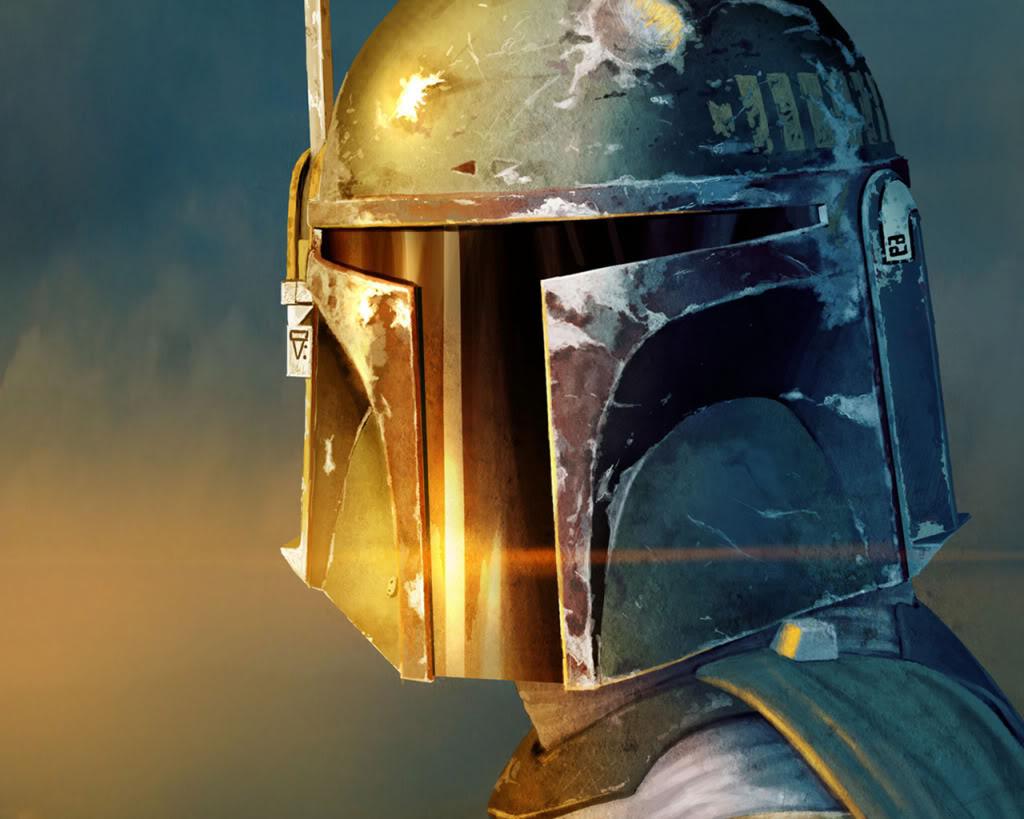}\end{minipage}  \quad \begin{minipage}[b]{.23\textwidth}\includegraphics[width=1\linewidth]{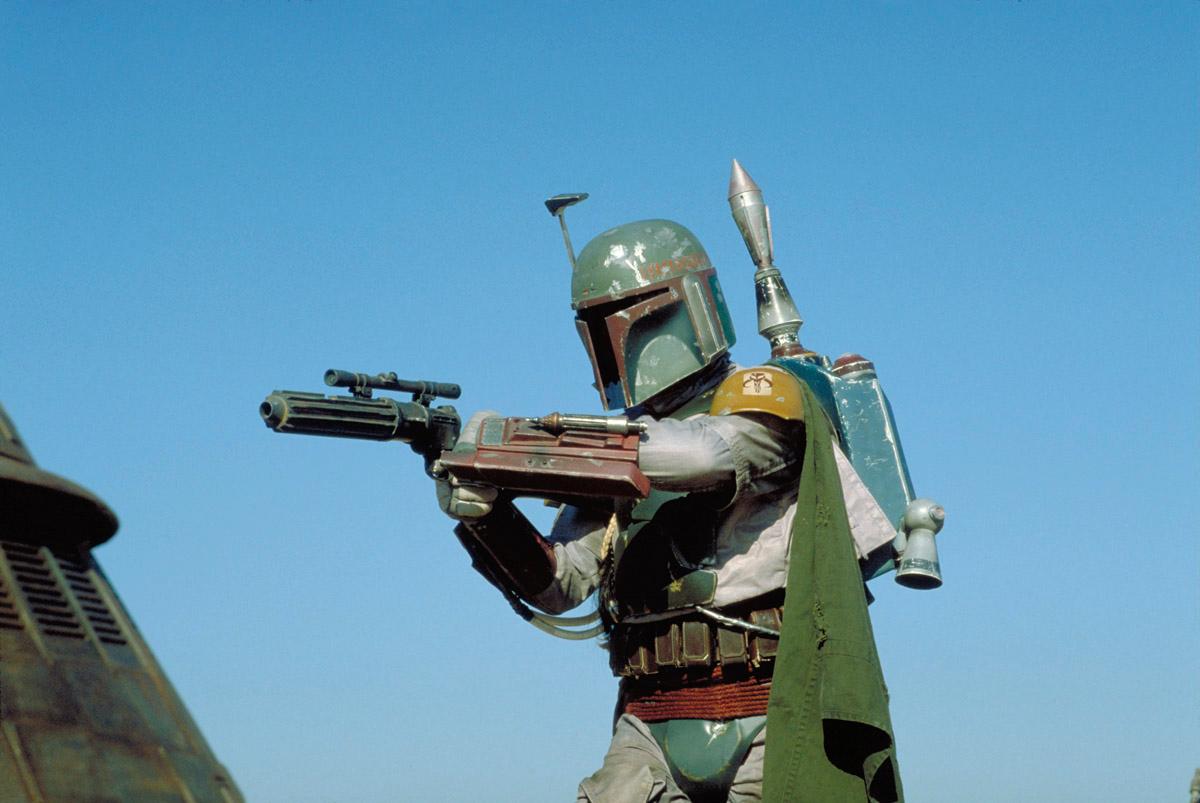}\end{minipage}\quad \begin{minipage}[b]{.23\textwidth}\includegraphics[width=1\linewidth]{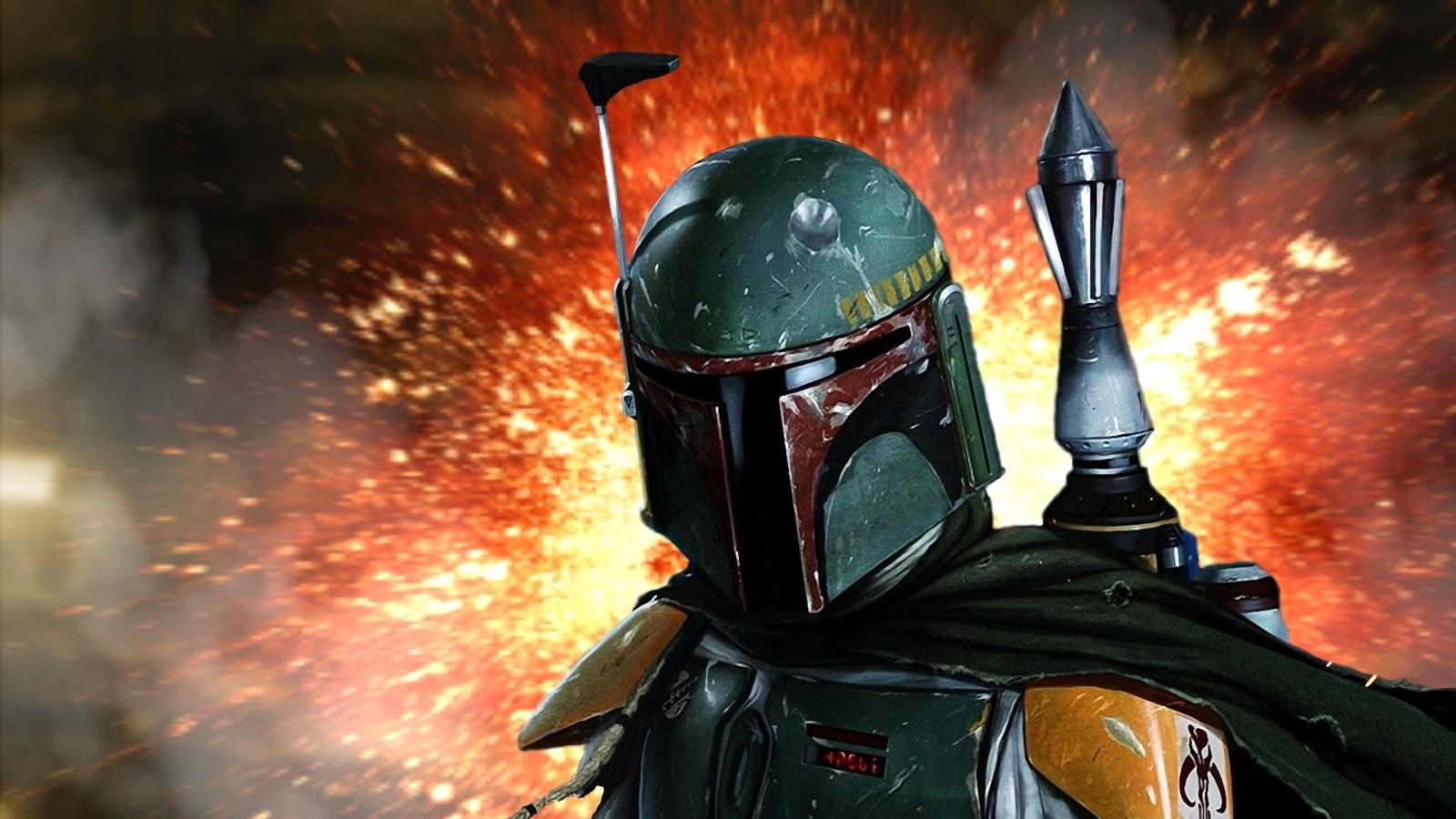}\end{minipage}\quad \qquad \qquad  \begin{minipage}[b]{.23\textwidth}\includegraphics[width=1\linewidth]{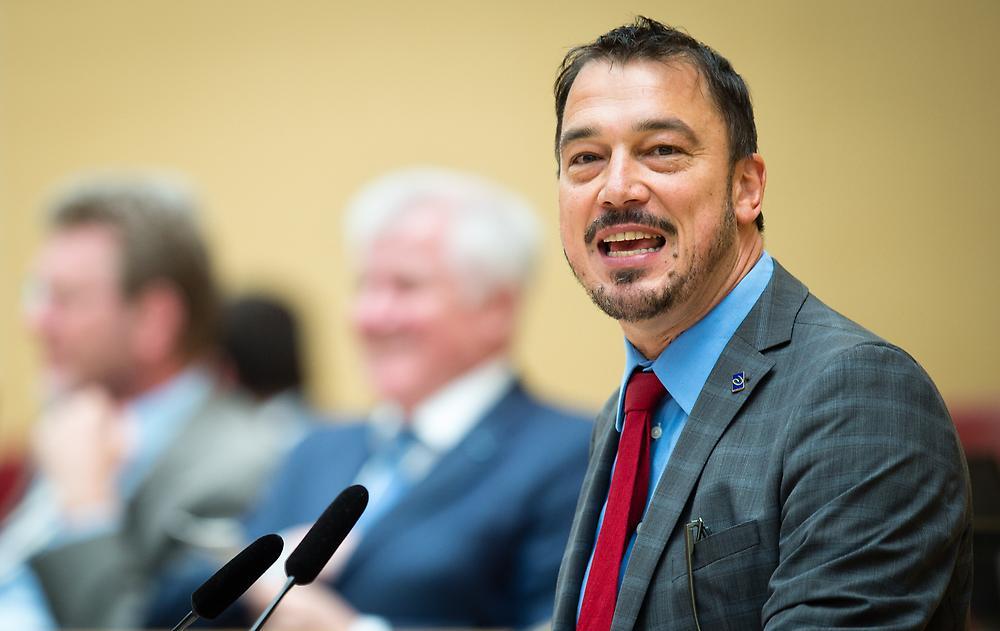}\end{minipage} \quad \begin{minipage}[b]{.23\textwidth}\includegraphics[width=1\linewidth]{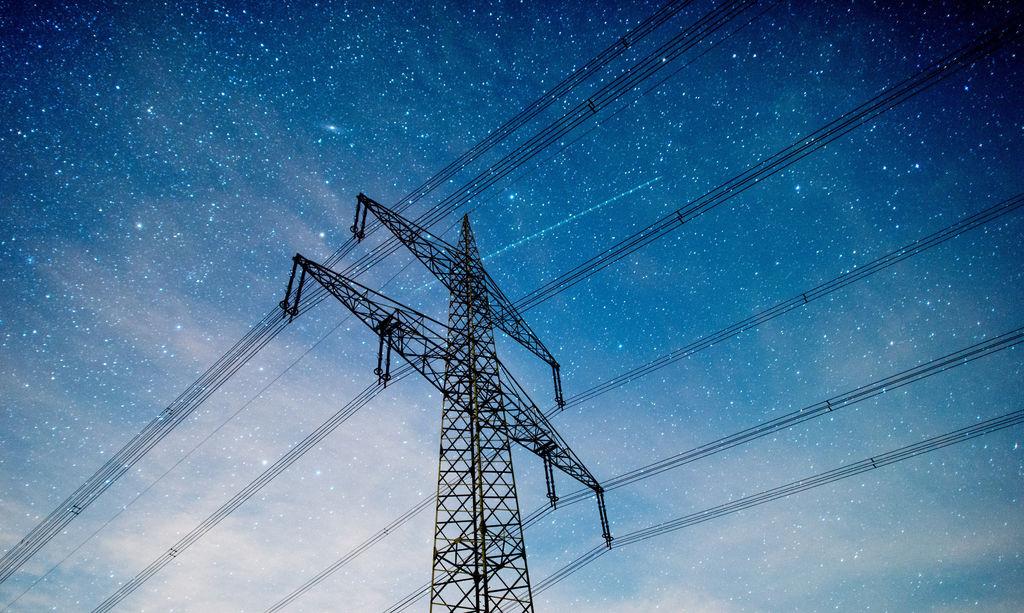}\end{minipage}  \quad \begin{minipage}[b]{.23\textwidth}\includegraphics[width=1\linewidth]{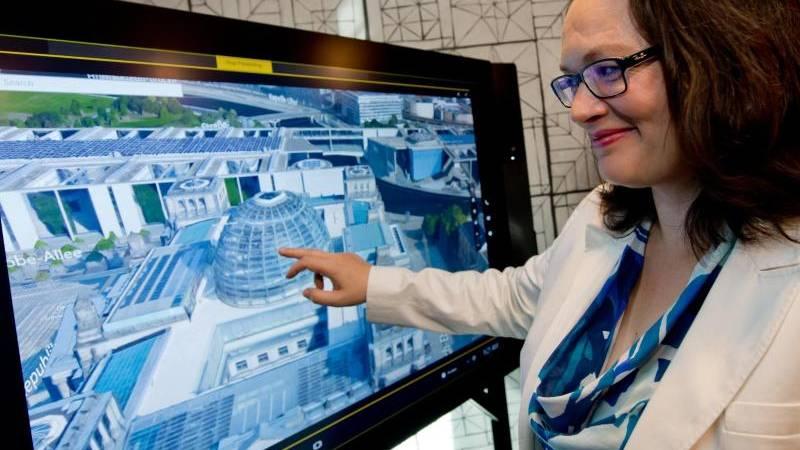}\end{minipage}\quad \begin{minipage}[b]{.23\textwidth}\includegraphics[width=1\linewidth]{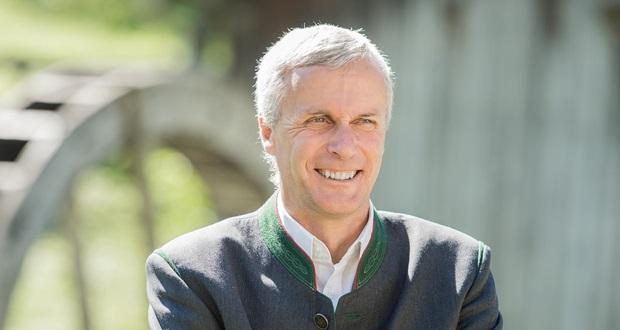}\end{minipage}} 
}
\caption{The first 4 images in the adjective image sets with lowest (left) and highest (right) dispersion values, which correspond to \textit{bold} and \textit{necessary} (first row), and the German translations of the English image words, \textit{fett} and \textit{notwendig} (second row).}\label{f:imageSetsAdjectives}
\end{figure*}

\begin{figure*}[h]
\centering
\resizebox{\textwidth}{!}{
\subfigure{
\begin{minipage}[b]{.23\textwidth}\includegraphics[width=1\linewidth]{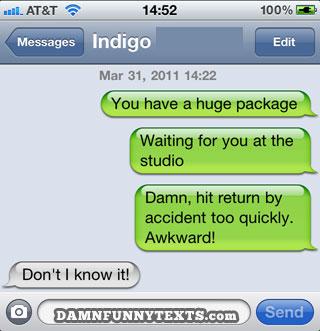}\end{minipage} \quad  \begin{minipage}[b]{.23\textwidth}\includegraphics[width=1\linewidth]{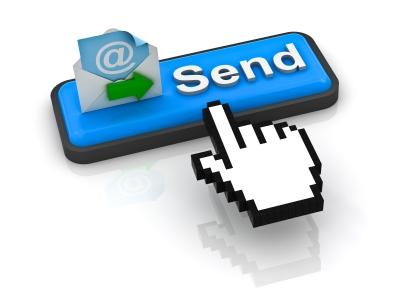}\end{minipage}\quad \begin{minipage}[b]{.23\textwidth}\includegraphics[width=1\linewidth]{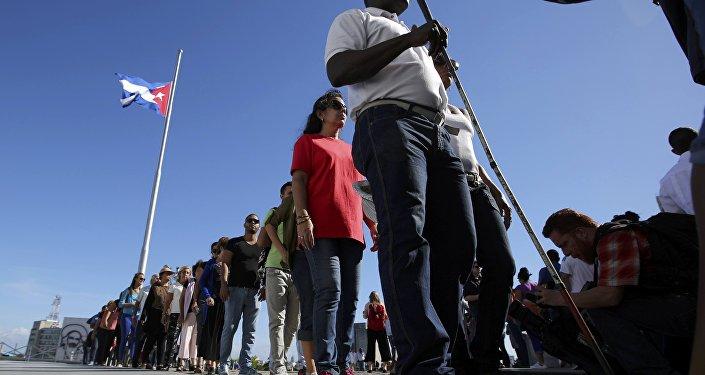}\end{minipage}\quad \begin{minipage}[b]{.23\textwidth}\includegraphics[width=1\linewidth]{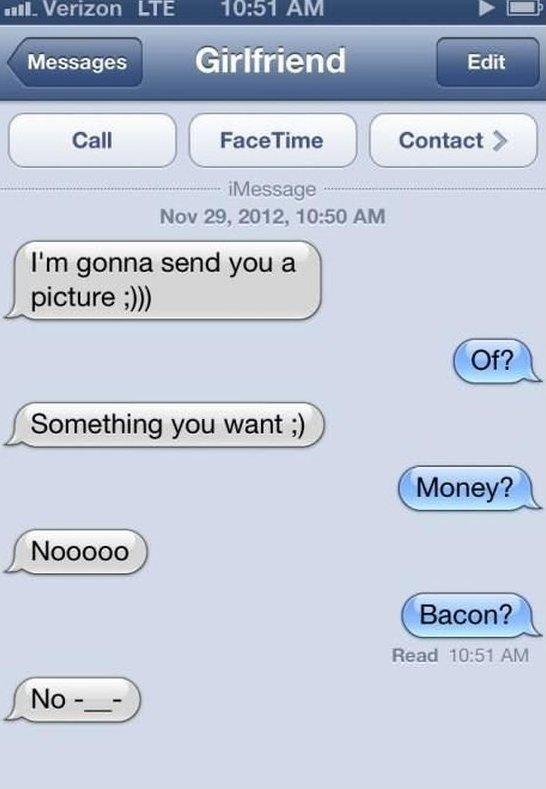}\end{minipage} \quad \qquad \qquad \begin{minipage}[b]{.23\textwidth}\includegraphics[width=1\linewidth]{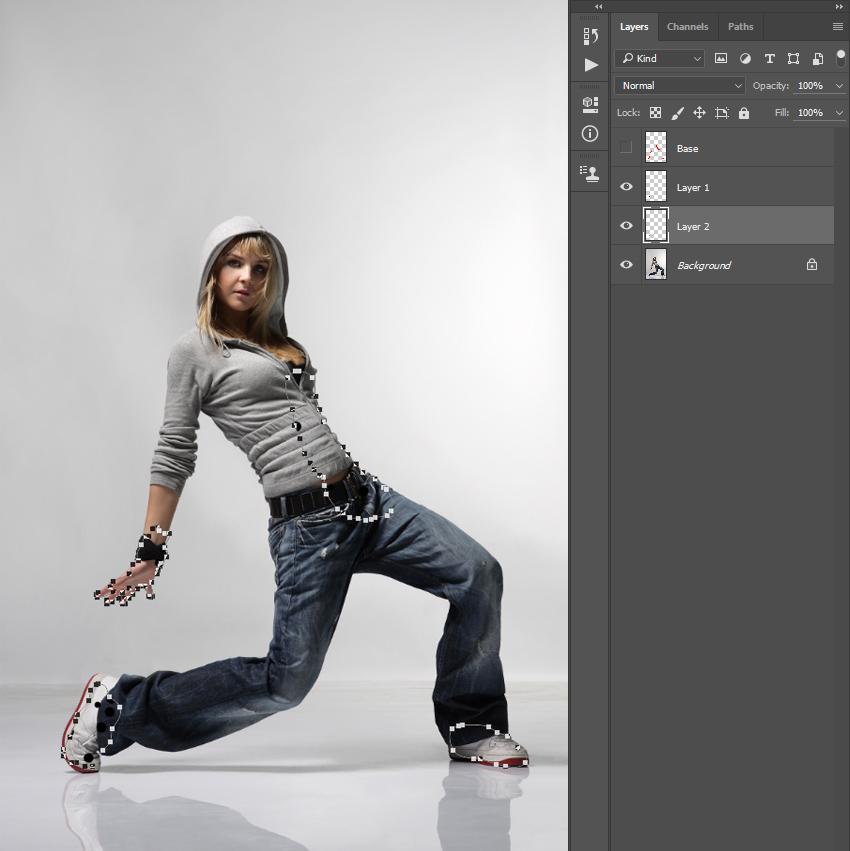}\end{minipage} \quad \begin{minipage}[b]{.23\textwidth}\includegraphics[width=1\linewidth]{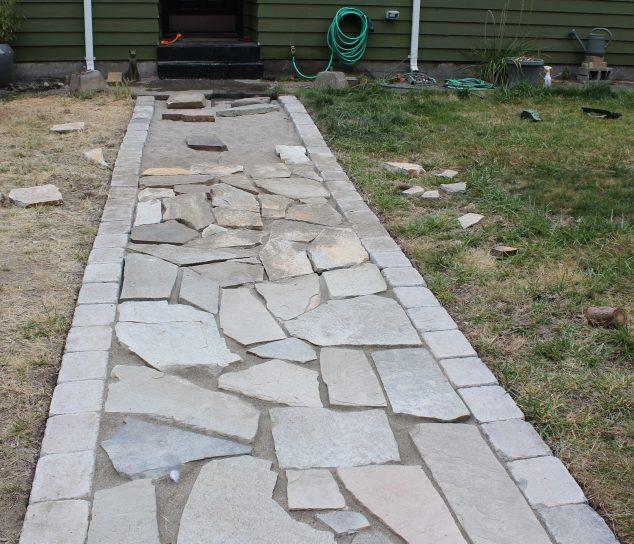}\end{minipage}  \quad  \begin{minipage}[b]{.23\textwidth}\includegraphics[width=1\linewidth]{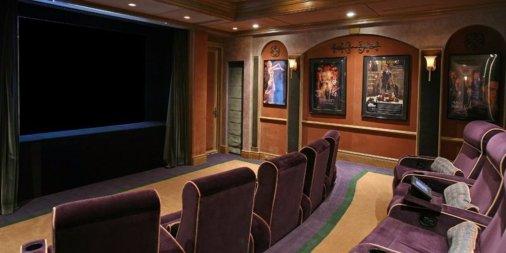}\end{minipage}\quad \begin{minipage}[b]{.23\textwidth}\includegraphics[width=1\linewidth]{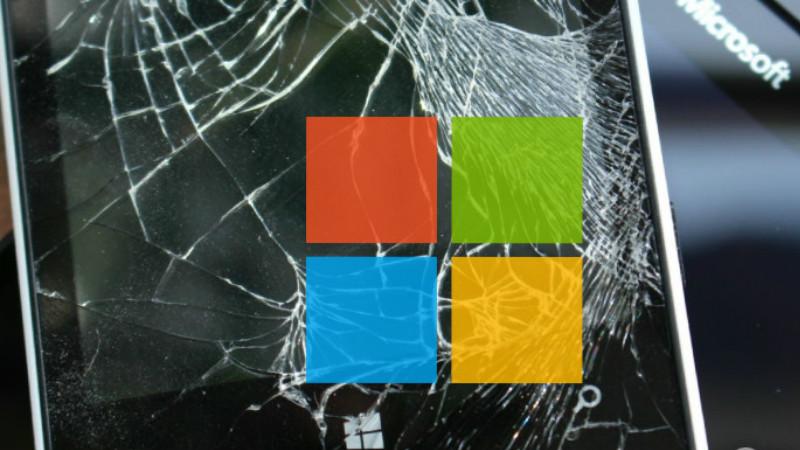}\end{minipage}} 
}
\resizebox{\textwidth}{!}{\subfigure{
\begin{minipage}[b]{.23\textwidth}\includegraphics[width=1\linewidth]{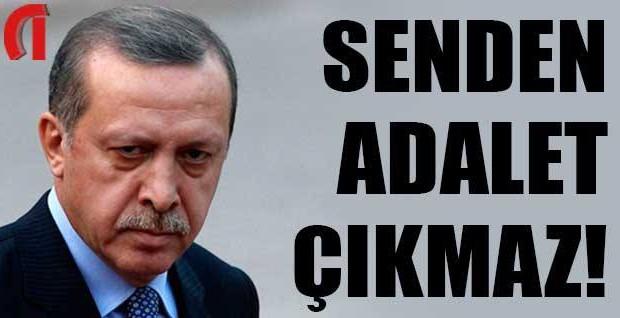}\end{minipage} \quad \begin{minipage}[b]{.23\textwidth}\includegraphics[width=1\linewidth]{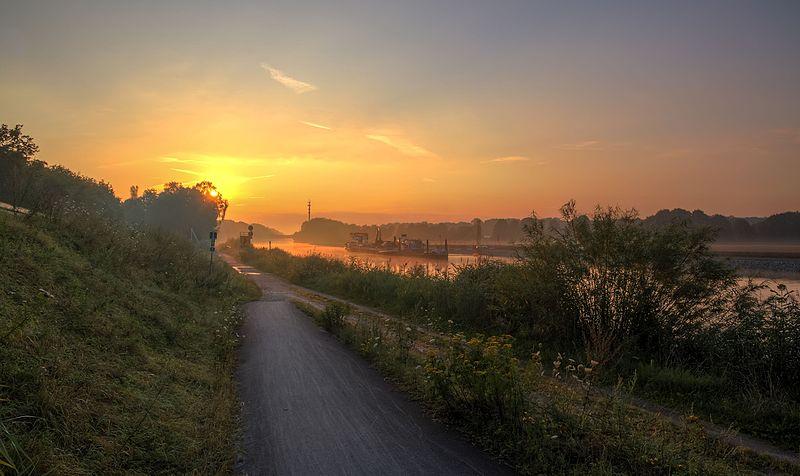}\end{minipage}  \quad  \begin{minipage}[b]{.23\textwidth}\includegraphics[width=1\linewidth]{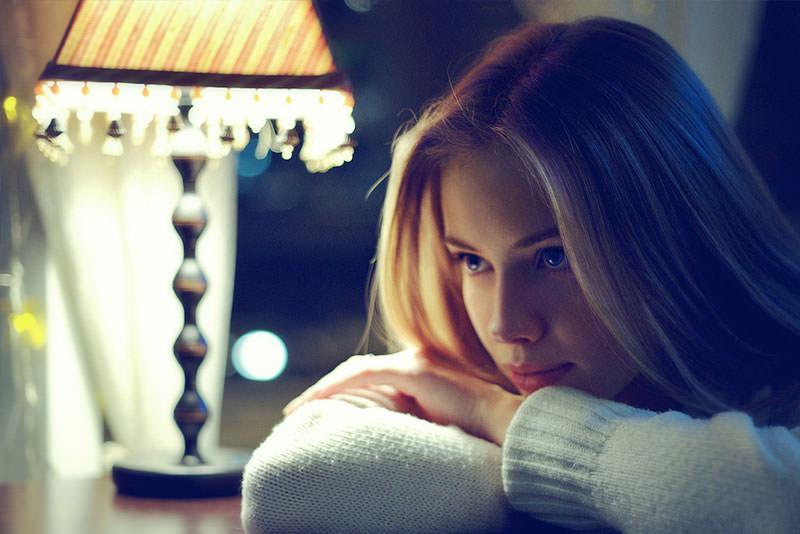}\end{minipage}\quad \begin{minipage}[b]{.23\textwidth}\includegraphics[width=1\linewidth]{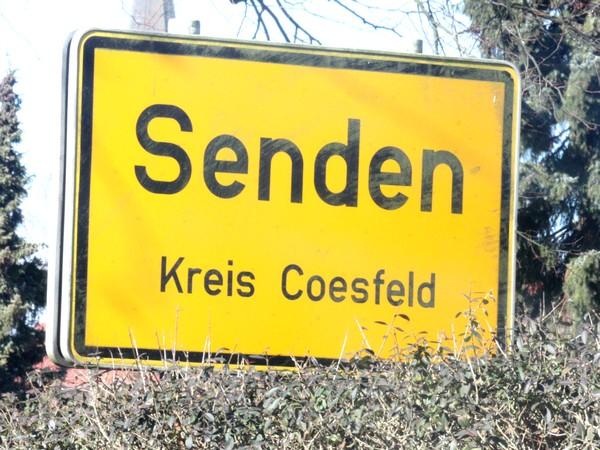}\end{minipage}\quad \qquad \qquad \begin{minipage}[b]{.23\textwidth}\includegraphics[width=1\linewidth]{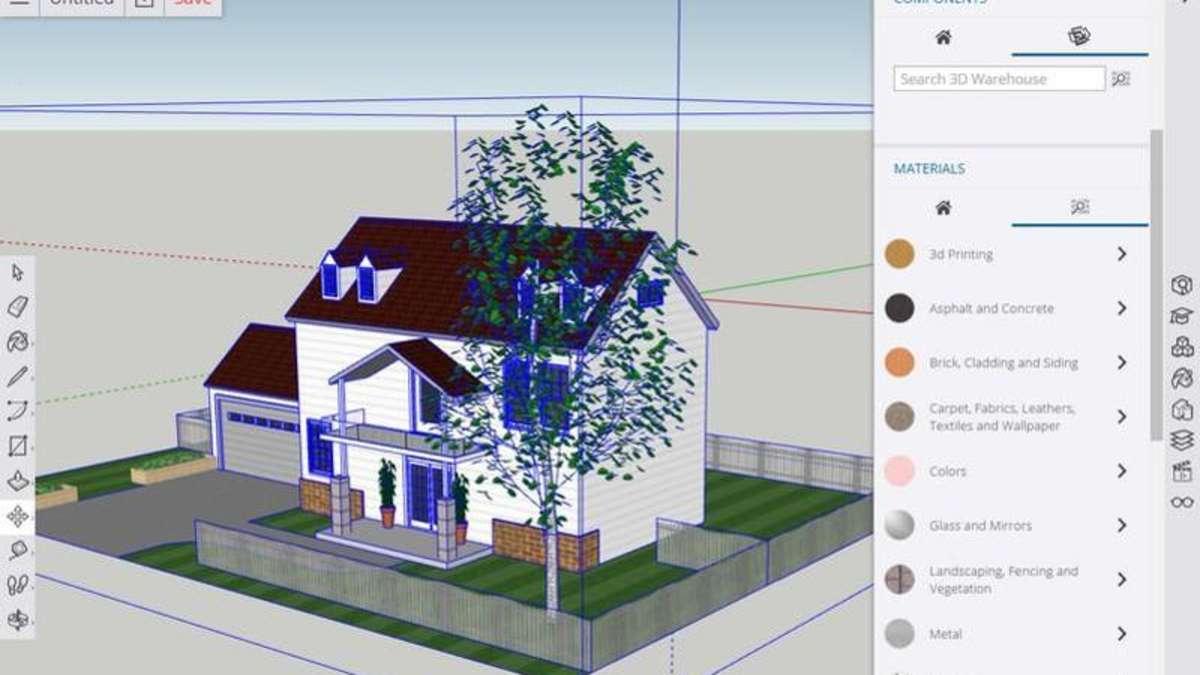}\end{minipage} \quad \begin{minipage}[b]{.23\textwidth}\includegraphics[width=1\linewidth]{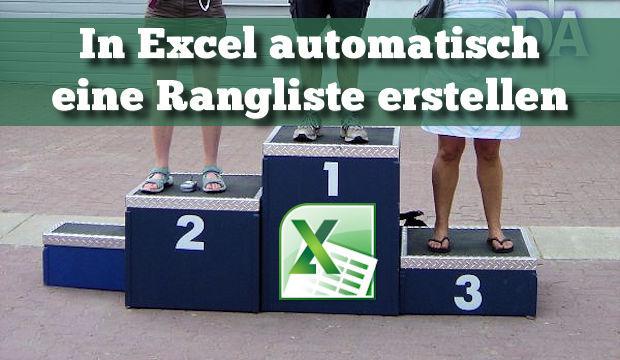}\end{minipage}\quad \begin{minipage}[b]{.23\textwidth}\includegraphics[width=1\linewidth]{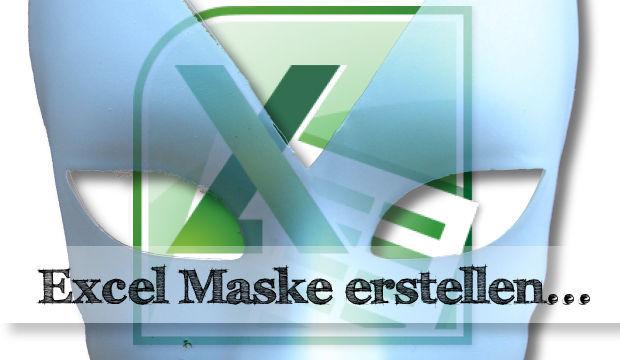}\end{minipage}\quad \begin{minipage}[b]{.23\textwidth}\includegraphics[width=1\linewidth]{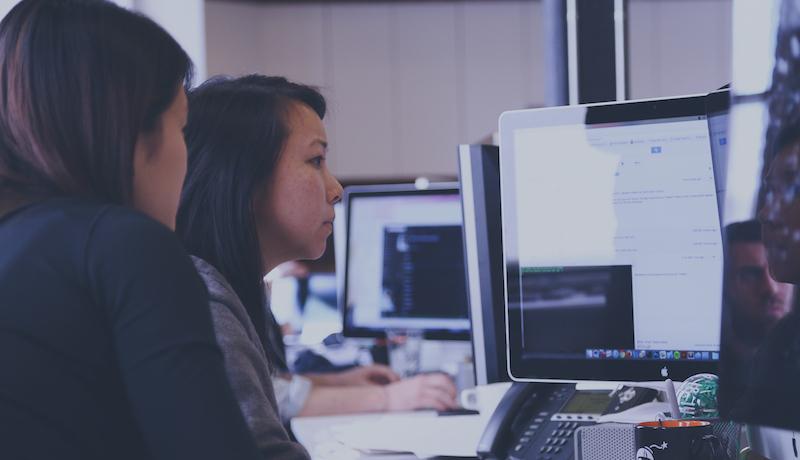}\end{minipage}} 
}
\caption{The first 4 images in the verb image sets with lowest (left) and highest (right) dispersion values, which correspond to \textit{send} and \textit{create} (first row), and the German translations of the English image words, \textit{senden} and \textit{erstellen} (second row).}\label{f:imageSetsVerbs}
\end{figure*}

We present examples of the image sets with the highest and the lowest dispersion values for each part-of-speech in Figures \ref{f:imageSetsNouns}, \ref{f:imageSetsVerbs} and \ref{f:imageSetsAdjectives}. Along with the image sets for English words, we also present examples from the images sets representing the German translations of the English words. Among the first 10 images returned by the search engine, we manually choose the 4 images that seem to be most interesting.\footnote{The 4 presented images are only a small part of the whole images set and cannot capture the whole variance in the images contained in a set. However, looking at these examples allows us to illustrate some weaknesses of our approach.}

\paragraph{Nouns}
For the nouns (Figure \ref{f:imageSetsNouns}), \textit{bicycle} is represented by the image set with the lowest dispersion. The images in both the English and the German image sets are highly similar, mostly showing iconic versions of bicycles that might be advertised for sale. This observation seems to apply for the other languages, too, as the correct translation for \textit{bicycle} is ranked on first position for all language pairs.

Considering the noun with highest dispersion, \textit{second}, we find that the images in the associated image set display at least two different senses of the word: its adjectival sense as in \textit{second position} and its sense as a fraction of a minute. The German translation \textit{Sekunde} mainly refers to the time unit. This is reflected in the associated images showing a clock and sports settings (where \textit{Sekunde} is associated with playing time).

\paragraph{Adjectives}
The example images for the adjectives (Figure \ref{f:imageSetsAdjectives}) are the most interesting, as they reveal some profound weaknesses of our approach. The adjective with lowest dispersion is \textit{bold}, which we would expect to be represented by images associated with its most frequent sense (according to WordNet\footnote{http://wordnet.princeton.edu/}: \textit{fearless}, \textit{daring}). However, the images in the \textit{bold} image set display a mobile phone, as one of the phones by the BlackBerry brand is named \textit{BlackBerry Bold}. In fact, 95\% of the images in the image set display a mobile phone. Here, the images provided by the search engine are clearly not associated with the adjectival meaning of \textit{bold}. 
The adjective \textit{bold} was translated by the translation system we use for dataset creation (see Section \ref{sec:data}) into the German adjective \textit{fett}, which is the translation of a very specific sense of \textit{bold} (as in \textit{bold face type}). The most frequent sense of the German \textit{fett} is \textit{fat} as in \textit{fat person}. Hence, we would expect to find some images of fat people or animals in the German image set. However, we find that 90\% of the images in the \textit{fett} image set display the Star Wars character \textit{Boba Fett} as shown in Figure \ref{f:imageSetsAdjectives}. Similar as it is the case for the English word \textit{bold}, the image search engine does not provide images associated with the adjectival sense of the queried word. For this word pair, even though the image dispersion is low, the model will fail to learn a correct translation based on the image data, as there is no semantic connection between the concepts in both image sets. 

For the adjective with highest dispersion, \textit{necessary}, we see various images, but at least some of them have an association with the meaning of the adjective. The same applies to the images for the German translation \textit{notwendig}. Based on these observations, we conclude that even if an image word might be displayed by images with a low dispersion, this does not imply that it is easy to learn a translation for it.


\paragraph{Verbs}

For the verbs (Figure \ref{f:imageSetsVerbs}), \textit{send} has the lowest dispersion (which is considerably higher than the dispersion for the noun with lowest dispersion, see Table \ref{t:exampleWords}). 44\% of the images in the \textit{send} image set show a smartphone chat like the one in images 1 and 4 in Figure \ref{f:imageSetsVerbs}. 

For the verb with highest dispersion, \textit{create}, we find a lot of images that have no obvious relation to the verb, like the third and fourth image on the right side of Figure \ref{f:imageSetsVerbs}. The German translation, \textit{erstellen}, is represented by many images that show the process of creating a document with a computer program. 

Clearly, we observe that most of the images returned by the search engine are displaying a concept that is associated with the verb, rather than the activity the verb refers to itself.

\subsection{Discussion}
The analysis described in the previous section gives some insights on why learning translations for adjectives and verbs from image search results works poorly. First, we observe that as soon as we move away from concrete nouns like \textit{bicycle}, the images returned by the search engine are diverse and the collected image sets have a high dispersion. The images often display concepts that are associated with the queried words rather than the concepts behind the words themselves. 

While this problem seems to be inherent to the process of learning representations of verbs and adjectives from image data in general, our analysis also reveals issues that are related to our specific approach, namely to rely on an image search engine to provide us with resources to learn from. In many cases, the search engine does not capture the intended sense of the query word, resulting in search results that are completely unrelated to the images of the query's translation. It is not clear how to query for verbs and adjectives and it is also not clear what mechanisms the search engine internally uses to group results. Even if we manage to enforce querying for a specific part-of-speech, the search engine might return images associated with an uncommon sense of that word. We depend on the search engine's interpretation of our query, which is especially problematic if the query not a concrete noun. We conclude that performance in the lexicon induction task might increase if we eliminate the dependence on the search engine by moving to a different resource annotated with semantic tags or natural language captions, such as Flickr or Wikipedia.

\section{Conclusion}
We showed that existing work on learning cross-lingual word representations from images obtained via web image search does not scale to other parts-of-speech than nouns. Even with supervised learning from image features, we are not able to obtain positive results for verbs and adjectives.
In our analysis, we identify the dependence on the image search engine as one of the main problems of the approach, when applied to adjectives and verbs. Hence, we plan to switch to a different resource to collect image data in future work.


\bibliography{emnlp2017}
\bibliographystyle{emnlp_natbib}

\end{document}